\documentclass{article}
\usepackage{amsmath,amsfonts,amssymb,mathtools}
\usepackage{graphicx} 
\usepackage[unicode=true,
 bookmarks=false,
 breaklinks=false,pdfborder={0 0 1},colorlinks=false]
 {hyperref}
\hypersetup{colorlinks,citecolor=blue,filecolor=blue,linkcolor=blue,urlcolor=blue}
\usepackage{multirow}
\usepackage{fullpage}
\usepackage{natbib}
\usepackage{xcolor}
\usepackage{tcolorbox}
\usepackage{appendix}
\usepackage{algorithm}
\usepackage[ruled, algo2e]{algorithm2e}

\newcommand{\ex}[2]{\mathbb{E}_{#1}\left[#2\right]}
\newcommand{\exlim}[2]{\mathop\mathbb{E}\limits_{#1}\left[#2\right]}
\newcommand{\argmin}{\mathop{\mathrm{argmin}}} 
\newcommand{\argmax}{\mathop{\mathrm{argmax}}} 
\newcommand{\prnbig}[1]{\big({#1}\big)} 
\newcommand{\innprod}[1]{\big\langle{#1} \big\rangle} 

\def\EE{{\mathbb E}}

\newcommand{\cM}{\mathcal{M}}
\newcommand{\cS}{\mathcal{S}}
\newcommand{\cA}{\mathcal{A}}

\newcommand{\cD}{\mathcal{D}}
\newcommand{\cL}{\mathcal{L}}
\newcommand{\cH}{\mathcal{H}}

\newcommand{\KL}[2]{\mathsf{KL}\prnbig{{#1}\,\|\,{#2}}}

\newcommand{\figref}[1]{Figure~\ref{#1}}

\usepackage[utf8]{inputenc} 
\usepackage[T1]{fontenc}    
\usepackage{hyperref}       
\usepackage{url}            
\usepackage{booktabs}       
\usepackage{amsfonts}       
\usepackage{nicefrac}       
\usepackage{microtype}      
\usepackage{xcolor,colortbl}         

\usepackage{mathtools}
\usepackage{amsmath}
 
\usepackage{graphicx,subcaption} 
\usepackage{enumitem}

\newtheorem{lemma}{Lemma}
\newtheorem{remark}{Remark}
\newtheorem{theorem}{Theorem}

\newtheorem{assumption}{Assumption}

\allowdisplaybreaks

\usepackage{xspace}

\makeatletter
\DeclareRobustCommand\onedot{\futurelet\@let@token\@onedot}
\def\@onedot{\ifx\@let@token.\else.\null\fi\xspace}

\def\eg{\emph{e.g}\onedot} 
\def\ie{\emph{i.e}\onedot}

\makeatother

\definecolor{blue_ppt}{rgb}{0,0.6,0.93}

\newcommand\newsubcap[1]{\phantomcaption%
       \caption*{\figurename~\thefigure(\thesubfigure): #1}}

\newcommand{\algabb}{VPO\xspace}

\newcommand{\dnorm}{\mathcal{\pi}_{\text{cal}}}
\newcommand{\piref}{\pi_{\text{ref}}}
\newcommand{\rgt}{r^{\star}}

\title{Value-Incentivized Preference Optimization: \\ A Unified Approach to Online and Offline RLHF}

\author{Shicong Cen\thanks{Carnegie Mellon University; emails: \texttt{\{shicongc,tongyang,yuejiec\}@andrew.cmu.edu}. }  \\
CMU  
\and Jincheng Mei\thanks{Google Research; emails: \texttt{\{jcmei,kgoshvadi,hadai,sherryy,schuurmans,bodai\}@google.com}.}  \\
Google 
\and Katayoon Goshvadi \\
Google 
\and Hanjun Dai \\
Google 
\and Tong Yang\footnotemark[1] \\
CMU 
\and Sherry Yang \\
Google 
\and Dale Schuurmans \\
Google 
\and Yuejie Chi\footnotemark[1] \\
CMU 
\and Bo Dai \\
Google}

\date{May 2024; Revised Feb. 2025}

\begin{document}
 
\maketitle

\begin{abstract}

Reinforcement learning from human feedback (RLHF) has demonstrated great promise in aligning large language models (LLMs) with human preference. Depending on the availability of preference data, both online and offline RLHF are active areas of investigation. A key bottleneck is understanding how to incorporate uncertainty estimation in the reward function learned from the preference data for RLHF, regardless of how the preference data is collected. While the principles of optimism or pessimism under uncertainty are well-established in standard reinforcement learning (RL), a practically-implementable and theoretically-grounded form amenable to large language models is not yet available, as standard techniques for constructing confidence intervals become  intractable under arbitrary policy parameterizations. 

In this paper, we introduce a unified approach to online and offline RLHF --- value-incentivized preference optimization (\algabb) --- which regularizes the maximum-likelihood estimate of the reward function with the corresponding value function, modulated by a {\em sign} to indicate whether the optimism or pessimism is chosen. \algabb also directly optimizes the policy with implicit reward modeling, and therefore shares a simpler RLHF pipeline similar to direct preference optimization. Theoretical guarantees of \algabb are provided for both online and offline settings, matching the rates of their standard RL counterparts. Moreover, experiments on text summarization and dialog verify the practicality
and effectiveness of \algabb.

\end{abstract}


\tableofcontents

\section{Introduction}

Fine-tuning large language models (LLMs) by {\em reinforcement learning from human feedback} (RLHF) \citep{ziegler2019fine} has been shown to significantly improve the helpfulness, truthfulness and controllability of LLMs, as illustrated by InstructGPT \citep{ouyang2022training} and many follow-ups. Roughly speaking, there are two critical components of RLHF: (1) {\em reward modeling}, which maps human preference rankings into a quantitative reward function that can guide policy improvement; and (2) {\em RL fine-tuning}, which seeks to adjust LLM output to align with human preferences by leveraging the learned reward function, i.e., increasing the probability of preferred answers and decreasing the probability of unfavored answers.  

Evidently, the curation of preference data is instrumental in the performance of RLHF, which is commonly modeled as pairwise comparisons from a Bradley-Terry ranking model \citep{bradley1952rank}.
In particular, given a query $x$, human annotators choose a preferred answer from two candidate answers $y_1$ and $y_2$ generated by an LLM. Despite the simple form, collecting large-scale and high-quality preference data can be expensive and time-consuming. Depending on the availability of preference data, two paradigms of RLHF are considered: (1) \emph{offline} RLHF, where only a pre-collected preference dataset is available, possibly generated from a pre-trained LLM after supervised fine-tuning (SFT); and (2) \emph{online} RLHF, where additional preference data can be collected adaptively to improve alignment. While initial work on RLHF focused on the offline setting, the online setting has also begun to receive considerable attention, as even a small amount of additional preference data has been shown to greatly boost performance.   

There has been significant work on the theoretical underpinnings of RLHF that seeks to uncover algorithmic improvements. Notably, while the original RLHF pipeline decouples reward modeling from RL fine-tuning, direct preference optimization (DPO) \citep{rafailov2023direct} integrates these as a single step in the {\em offline} setting, leveraging a closed-form solution for the optimal policy in the RL fine-tuning phase. This has led to a welcome simplification of the RLHF pipeline,
allowing direct optimization of the policy (i.e., the LLM) from preference data.

Nevertheless, significant challenges remain in RLHF, particularly concerning how to incorporate estimates of reward {\em uncertainty} in direct preference optimization when parameterizing policies with large-scale neural networks --- such as LLMs --- in a theoretically and practically effective manner. In standard reinforcement learning (RL), managing uncertainty when an agent interacts with an environment is a critical aspect in achieving near-optimal performance \citep{sutton2018reinforcement}, when using methods that range from policy-based \citep{schulman2017proximal,xiao2021optimality}, value-based \citep{mnih2015human,kumar2020conservative}, and actor-critic methods \citep{mnih2016asynchronous}. 
One dominant approach in the bandit setting, for example, is to construct confidence intervals of the reward estimates, then acting according to the upper and lower confidence bounds --- following the principles of optimism and pessimism in the online and offline settings respectively \citep{lattimore2020bandit,lai1985asymptotically,rashidinejad2022bridging}.

Despite the fact that uncertainty estimation is even more critical in RLHF, due to the coarse nature of preference data, effective implementations of theoretically justified optimistic and pessimistic principles have yet to be developed in the RLHF literature. For example, existing online preference alignment methods, such as Nash-MD \citep{munos2023nash} and OAIF \citep{guo2024direct}, do not incorporate exploration; similarly,  pessimism is also not implemented in offline preference alignment methods, such as DPO \citep{rafailov2023direct} and IPO \citep{azar2024general}. A key reason for these omissions is that it is extremely difficult to construct confidence intervals for arbitrary neural networks \citep{gawlikowski2021survey}, let alone LLMs. Since optimism for online exploration and pessimism for offline RL both require uncertainty estimation, and given the difficulty of conducting uncertainty estimation for large-scale neural networks, a natural and important question arises: 
\begin{center}
\emph{Can we implement the optimistic/pessimistic principles  
under uncertainty 
in a practically efficient manner for online/offline preference alignment in LLMs while retaining theoretical guarantees?}
\end{center}

\subsection{Our contributions}

In this paper, we provide affirmative answer to the question. Our major contributions are as follows.  
\begin{itemize}
    \item[\textbf{(i)}] We propose value-incentivized preference optimization (\algabb) for both online and offline RLHF, a unified algorithmic framework that {\em directly optimizes the LLM policy} with the optimistic/pessimistic principles under uncertainty. Avoiding explicit uncertainty estimation, \algabb regularizes maximum likelihood estimation of the reward function toward (resp. against) responses that lead to the highest value in the online (resp. offline) setting, hence implementing optimism (resp. pessimism). Theoretical regret guarantees of \algabb are developed for both  online and offline RLHF, matching their corresponding rates in the standard RL literature with explicit uncertainty estimation. 

    \item[\textbf{(ii)}]   In addition, \algabb reveals the critical role of reward calibration, where the shift ambiguity of the reward model inherent in the Bradley-Terry model \citep{bradley1952rank} can be exploited to implement additional behavior regularization \citep{pal2024smaug,ethayarajh2024kto} via centering the reward model with respect to a {\em calibration policy}. This allows \algabb to provide a theoretical foundation for popular conservative offline RL methods (e.g., \citep{kumar2020conservative}), as well as regularized RLHF methods (e.g., DPOP \citep{pal2024smaug}).

    \item[\textbf{(iii)}] \algabb admits a practically-implementable form suitable for RLHF on LLMs, and more generally, deep-learning architectures. We conduct extensive experimental studies using \texttt{TL;DR} and \texttt{ARC-Challenge} tasks in online and offline settings with optimistic and pessimistic bias, respectively.  
    The results demonstrate improved empirical performance. 

\end{itemize}

\subsection{Related work}

\paragraph{RLHF.} Since the introduction of the original RLHF framework, there have been many proposed simplifications of the preference alignment procedure and attempts to improve performance, including but not limited to SLiC \citep{zhao2023slic}, GSHF \citep{xiong2023gibbs}, DPO \citep{rafailov2023direct}, and its variants, such as Nash-MD \citep{munos2023nash}, IPO \citep{azar2024general}, OAIF \citep{guo2024direct}, SPO \citep{swamy2024minimaximalist}, SPIN \citep{chen2024self}, WIND \citep{yang2024faster}, GPO \citep{tang2024generalized}, and DPOP \citep{pal2024smaug}. These methods can roughly be grouped into online and offline variants, depending on whether preference data is collected before training (offline) or by using the current policy during training (online).  

In offline preference alignment, identity preference optimization (IPO, \citep{azar2024general}) argues that it is problematic to use the Bradley-Terry model in DPO to convert pairwise preferences into pointwise reward values, and proposes an alternative objective function to bypass the use of the Bradley-Terry model. DPO-Positive (DPOP, \citep{pal2024smaug}) observes a failure mode of DPO that the standard DPO loss can reduce the model’s likelihood on preferred answers, and proposes to add a regularization term to the DPO objective to avoid such a failure mode.
On the other hand, online AI
feedback (OAIF, \citep{guo2024direct}) proposes an online version of DPO, where online preference data from LLM annotators is used to evaluate and update the current LLM policy in an iterative manner. Iterative reasoning preference optimization (Iterative RPO, \citet{pang2024iterative}) proposes to add an additional negative log-likelihood term in the DPO loss to improve performances on reasoning tasks. Finally, \citet{chang2024dataset} proposes to reuse the offline preference data via reset.

\paragraph{Uncertainty estimation in RL.} The principles of optimism and pessimism are typically implemented via constructing confidence intervals or posterior sampling, 
which have been demonstrated to be provably efficient in tabular settings \citep{jin2018q,shi2022pessimistic}. Yet, these approaches have had limited success in conjunction with deep learning architectures \citep{gawlikowski2021survey}, and many empirical heuristics in turn lack theoretical validation \citep{kumar2020conservative}. Notwithstanding, alternative regularization schemes are developed for general function approximation settings with theoretical guarantees, such as Bellman-consistent pessimism for offline RL \citep{xie2021bellman}, and reward-biased exploration for online RL \citep{mete2021reward, liu2024maximize} and Markov games \citep{liu2024maximize,yang2025incentivize}. 
\algabb draws inspiration from reward-biased exploration \citep{kumar1982new,liu2019exploration,liu2024maximize,hung2021reward,mete2021reward} in the standard online RL literature, but significantly broadens its scope to the offline setting and RLHF for the first time.   

\paragraph{Concurrent work.} Since posting the initial version of this paper on arXiv, we discovered several concurrent work that also appeared online around the same time proposing similar regularization techniques as ours to encourage optimism (resp. pessimism) for online (resp. offline) RLHF \citep{zhang2024selfexploring,xie2024exploratory,liu2024provably}. Despite slightly different forms, the algorithms studied in \citet{zhang2024selfexploring,xie2024exploratory,liu2024provably} can be interpreted as adopting different choices of the {\em calibration policy} in \algabb. In the context of online RLHF, \citet{zhang2024selfexploring} empirically studied a similar algorithm as the proposed online \algabb under the contextual bandit formulation of RLHF; \citet{xie2024exploratory} provided finite-time regret analysis of a similar algorithm for the token-level MDP formulation with general function approximation, which extends to general deterministic contextual MDPs as well.
In the context of offline RLHF, \citet{liu2024provably} studied a similar algorithm as the proposed offline \algabb and provided a sample complexity analysis under the contextual bandit formulation, yet focusing on general function approximation and different assumptions.

\section{Preliminaries}

In RLHF, a language model is described by a policy $\pi$, which generates an answer $y \in \mathcal{Y}$ given prompt $x\in \mathcal{X}$ according to the conditional probability distribution $\pi(\cdot | x)$. The standard RLHF process consists of four stages: supervised fine-tuning (SFT), preference data generation, reward modeling, and RL fine-tuning. In the SFT stage, a language model $\pi_{\text{sft}}$ is obtained by fine-tuning a pre-trained LLM with supervised learning. The remaining stages continue training by leveraging the preference data, which we elaborate below.
 
\paragraph{Reward modeling from preference data.} An oracle (e.g., a human labeler or a scoring model) evaluates the quality of two answers $y_1$ and $y_2$ given prompt $x$ and reveals its preference. A widely used approach for modelling the probability of pairwise preferences is the Bradley–Terry model \citep{bradley1952rank}:
\begin{align} \label{eq:BT}
    \mathbb{P}(y_1\succ y_2|x) &= \frac{\exp(r^\star(x, y_1))}{\exp(r^\star(x, y_1))+\exp(r^\star(x, y_2))} = \sigma(r^\star(x, y_1) - r^\star(x, y_2)),
\end{align}
where $y_1\succ y_2$ indicates that $y_1$ is preferred over $y_2$, $r^\star: \mathcal{X}\times\mathcal{Y}\to \mathbb{R}$ is the ground truth reward function, and $\sigma : \mathbb{R} \to (0, 1)$ is the logistic function. A preference data sample is denoted by a tuple $(x, y_+, y_-)$, where $y_+$ (resp.  $y_{-}$) is the preferred (resp. unpreferred) answer in the comparison.

Given a preference dataset $\mathcal{D} = \{(x^i, y_+^i, y_-^i)\}$ composed of independent samples, the reward function $r$ can be estimated by maximum likelihood estimation (MLE):
\begin{equation}
    r_{\mathsf{MLE}} = \arg\min_{r} \; \ell(r, \mathcal{D}),
    \label{eq:MLE}
\end{equation}
where $\ell(r, \mathcal{D})$ is the negative log-likelihood of $\mathcal{D}$, given as
\begin{equation}
     \ell(r, \mathcal{D}) \coloneqq -\sum_{(x^i, y_+^i, y_-^i)\in\mathcal{D}}{\log \sigma(r(x^i,y_+^i) - r(x^i,y_-^i))}.
\end{equation}

\paragraph{RL fine-tuning.} Given a reward model $r$, we seek to fine-tune the policy $\pi$ to achieve an ideal balance between the expected reward and its distance from an initial policy $\pi_{\text{ref}}$, which is typically the same as $\pi_{\text{sft}}$. This is  achieved by maximizing the KL-regularized value function $J(r, \pi)$, defined as
\begin{equation} \label{eq:KL_reward}
    J(r, \pi) = \exlim{x \sim \rho, y \sim \pi(\cdot|x)}{r(x, y) } - \beta\exlim{x \sim \rho}{ \KL{\pi(\cdot|x)}{\pi_{\text{ref}}(\cdot|x)}},
\end{equation}
where $\KL{\pi_1}{\pi_2} $ is the KL divergence from $\pi_1$ to $\pi_2$, and $\beta>0$ is a regularization parameter. Consequently, the RL fine-tuned policy $\pi_r$ with respect to the reward $r$ satisfies  
\begin{equation}
    \pi_r \coloneqq \arg\max_{\pi} J(r, \pi),
\end{equation}
which admits a closed-form solution \citep{rafailov2023direct}, i.e.,
\begin{equation}
\forall (x\times y) \in \mathcal{X}\times\mathcal{Y}: \qquad    \pi_r(y|x) = \frac{\pi_{\text{ref}}(y|x)\exp(r(x,y)/\beta)}{Z(r, x)}.
    \label{eq:RLHF-policy-closed-form}
\end{equation}
Here, $ Z(r, x)$ is a normalization factor given by
\begin{equation} \label{eq:normalization}
    Z(r, x) = \sum_{y' \in \mathcal{Y}}\pi_{\text{ref}}(y'|x)\exp(r(x,y')/\beta).
\end{equation}

\paragraph{Direct preference optimization.} 
The closed-form solution \eqref{eq:RLHF-policy-closed-form} allows us to write the reward function $r$ in turn as
\begin{equation}\label{eq:reward-policy-eq}
    r(x, y) = \beta (\log \pi_r(y|x) - \log \pi_{\text{ref}}(y|x) + \log Z(r, x)).
\end{equation}
Plugging the above equation into the reward MLE~\eqref{eq:MLE}, we obtain the seminal formulation of direct preference optimization (DPO) over the policy space \citep{rafailov2023direct}, 
\begin{equation} \label{eq:DPO}
    \pi_{\mathsf{DPO}} = \arg\min_{\pi} \; -\sum_{(x^i, y_+^i, y_-^i)\in\mathcal{D}}{\log \sigma\left( \beta \left(\log \frac{\pi(y_+^i|x)}{\pi_{\text{ref}}(y_+^i|x)}  - \log \frac{\pi(y_-^i|x)}{\pi_{\text{ref}}(y_-^i|x)}  \right) \right)} ,
\end{equation}
which avoids explicitly learning the reward model.

\section{Value-Incentivized Preference Optimization}

A major caveat of the standard RLHF framework concerns the lack of accounting for reward uncertainty, which is known to be indispensable in the success of standard RL paradigms in both online and offline settings \citep{cesa2017boltzmann,rashidinejad2022bridging}. This motivates us to investigate a principled mechanism that be easily integrated into the RLHF pipeline, while bypassing the difficulties of explicit uncertainty estimation in LLMs.

\subsection{General framework}

In view of the sub-optimality of naive MLE for reward estimation \citep{cesa2017boltzmann,rashidinejad2022bridging}, and motivated by the effectiveness of reward-biased MLE in online RL \citep{kumar1982new,liu2019exploration,liu2024maximize}, we propose to regularize the reward estimate via
\begin{equation} 
J^\star(r) = \max_{\pi} \, J(r,\pi),
\end{equation}
which measures the resulting value function for the given reward if one acts according to its optimal policy. However, in RLHF,  by the definition \eqref{eq:BT}, the reward function $r^\star$ is only identifiable up to a prompt-dependent global shift. Specifically, letting $r_1(x, y) = r_2(x, y) + c(x)$ be two reward functions that only differ by a prompt-dependent shift $c(x)$, we have $r_1(x, y_1) - r_1(x, y_2) = r_2(x, y_1) - r_2(x, y_2)$, which leads to $J^{\star}(r_1) = J^{\star}(r_2) +\mathbb{E}_{x\sim \rho}[c(x)]$. To resolve this challenge, we introduce the following equivalent class of reward functions for the Bradley-Terry model to eliminate the shift ambiguity, which also has the calibration effect of centering the reward function while offering a regularization mechanism to incorporate additional policy  preferences.

\begin{assumption}
\label{assump:zero_mean}
We assume that $r^\star \in\mathcal{R}$, where  
\begin{equation}
    \mathcal{R} = \bigg\{r:
    \exlim{\substack{x\sim\rho, y \sim \dnorm(\cdot| x)}}{r(x, y)} = 0.
    \bigg\}.
    \label{eq:zero_pi_mean}
\end{equation}
Here, $\rho$ is the prompt distribution and $\dnorm$ is a fixed calibration distribution independent of the algorithm.
\end{assumption}

The proposed regularized MLE of the Bradley-Terry model \eqref{eq:MLE}  
appends a  bias term to the negative likelihood  
\begin{equation}
    r_{\mathsf{VPO}} = \arg\min_{r \in \mathcal{R} } \; \{\ell(r, \mathcal{D}) - \mathsf{sign} \cdot \alpha \cdot J^\star(r)\},
        \label{eq:RBMLE}
\end{equation}
incentivizing the algorithm to favor (resp. avoid) reward models with higher value  $J^\star(r) $ in the online (resp. offline) setting. Here, $\alpha > 0$ is a constant controlling the strength of regularization, and $\mathsf{sign}$ is set to $1$ in the online setting and $-1$ in the offline setting. 

At first glance, the objective function for \algabb \eqref{eq:RBMLE} does not immediately imply a computationally-efficient algorithm due to the presence of $J^{\star}(r)$. However, by exploiting the same closed-form solution for the optimal policy given the reward in~\eqref{eq:RLHF-policy-closed-form},  
and the reward representation inferred from the policy via \eqref{eq:reward-policy-eq}, 
we can explicitly express $J^\star(r)$ as 
\begin{align}
    J^\star(r) &= \exlim{x \sim \rho, y \sim \pi_r(\cdot|x)}{r(x, y) - \beta (\log \pi_r(y|x) - \log\pi_{\text{ref}}(y|x))}\notag\\
    &= \exlim{x \sim \rho, y \sim \pi_r(\cdot|x)}{\log Z(r, x)}\notag\\
    &= \exlim{x \sim \rho, y \sim \dnorm(\cdot|x)}{\log Z(r, x)}\notag\\
    &= \exlim{x \sim \rho, y \sim \dnorm(\cdot|x)}{r(x, y) - \beta (\log \pi_r(y|x) - \log\pi_{\text{ref}}(y|x))}\notag\\
    &= -\beta \exlim{x \sim \rho, y \sim \dnorm(\cdot|x)}{ \log \pi_r(y|x) - \log\pi_{\text{ref}}(y|x)}, \label{eq:Jstar}
\end{align}
where the second step follows because the bracketed term is independent of $y$ (c.f. \eqref{eq:RLHF-policy-closed-form}) and the last step follows from \eqref{eq:zero_pi_mean} whenever $r\in\mathcal{R}$.
Given this key ingredient, we can then rewrite \eqref{eq:RBMLE} to directly optimize the LLM policy, in a flavor similar to DPO, as
\begin{align}
    \pi_{\mathsf{VPO}} &= \argmin_{\pi_r:\, r\in \mathcal{R} } \, \{\ell(r, \mathcal{D}) -  \mathsf{sign} \cdot \alpha \cdot J^\star(r)\}\notag\\
   &= \argmin_{\pi_r:\, r\in\mathcal{R} } \Big\{-\sum_{(x^i, y_+^i, y_-^i)\in\mathcal{D}} {\log \sigma\Big(\beta\log\frac{\pi_r(y_+^{i} |x^{i})}{\pi_{\text{ref}}(y_+^{i}|x^{i})} - \beta\log\frac{\pi_r(y_-^{i}|x^{i})}{\pi_{\text{ref}}(y_-^{i}|x^{i})}\Big)} \notag\\
    &\qquad\qquad\qquad + \mathsf{sign} \cdot\alpha \beta  \exlim{x \sim \rho, y \sim \dnorm(\cdot|x)}{\log \pi_r(y|x) - \log\pi_{\text{ref}}(y|x)}\Big\} \nonumber \\
   &= \arg\min_{\pi} \Big\{-\sum_{(x^i, y_+^i, y_-^i)\in\mathcal{D}} {\log \sigma\Big(\beta\log\frac{\pi(y_+^{i} |x^{i})}{\pi_{\text{ref}}(y_+^{i}|x^{i})} - \beta\log\frac{\pi(y_-^{i}|x^{i})}{\pi_{\text{ref}}(y_-^{i}|x^{i})}\Big)} \notag\\
    &\qquad\qquad\qquad + \mathsf{sign} \cdot\alpha \beta  \exlim{x \sim \rho, y \sim \dnorm(\cdot|x)}{\log \pi (y|x) - \log\pi_{\text{ref}}(y|x)}\Big\},
    \label{eq:policy_RBMLE}
\end{align}
where we drop the constraint on $r\in \mathcal{R}$, since for any policy $\pi$ there exists $r\in\mathcal{R}$ such that $\pi =\pi_r$. 

Observing that the reference policy $\pi_{\text{ref}}(y|x)$ in the last term of \eqref{eq:policy_RBMLE} $\exlim{x \sim \rho, y \sim \dnorm(\cdot|x)}{\log \pi (y|x) - \log\pi_{\text{ref}}(y|x)} $ does not impact the optimization solution, we can change it to $\exlim{x \sim \rho, y \sim \dnorm(\cdot|x)}{\log \pi (y|x) - \log\dnorm(y|x)}\Big\} = - \exlim{x \sim \rho}{ \KL{\dnorm(\cdot|x)}{\pi(\cdot|x)}} $, which amounts to adding a  KL regularization to the original DPO, and offers an interesting interpretation as pushing $\pi$ against/towards $\dnorm$ in the online/offline settings respectively, unveiling the role of reward calibration in RLHF. 

In what follows, we elaborate the development of \algabb in both the online and offline settings with  corresponding theoretical guarantees under linear function approximation.

\subsection{Online RLHF: algorithm and theory}
The online RLHF procedure extends training by performing reward learning and policy learning iteratively, with a growing preference dataset collected by using the current policy. We use $\pi^{(t)}$ to denote the policy used in the $t$-th iteration, where the superscript $^{(t)}$ indicates iteration $t$ in the online setting.
 The $t$-th iteration of \algabb for online RLHF  consists of the following steps:
\begin{enumerate}
    \item \textbf{New preference data generation.} We sample a new prompt $x^{(t)}\sim\rho$ and two answers $y_1^{(t)}, y_2^{(t)}\sim\pi^{(t)}(\cdot|x^{(t)})$, query the preference oracle and append $(x^{(t)}, y_+^{(t)}, y_-^{(t)})$ to the preference dataset.
    \item \textbf{Reward learning.} We train a reward model with preference data $\mathcal{D}^{(t)} \coloneqq \{(x^{(s)}, y_+^{(s)}, y_-^{(s)})\}_{s=1}^{t}$ by minimizing the regularized negative log-likelihood, i.e., 
    \begin{equation}
        r^{(t+1)} = 
        \arg\min_{r\in\mathcal{R}} \; \{\ell(r, \mathcal{D}^{(t)}) - \alpha \cdot J^\star(r)\}. 
        \label{eq:RBMLE_online}
    \end{equation}
    \item \textbf{Policy learning.} This step trains the policy by solving the RL fine-tuning problem:
    \begin{equation}
        \pi^{(t+1)} = \arg\max_{\pi} J(r^{(t+1)}, \pi).
        \label{eq:RLHF-policy}
    \end{equation}
\end{enumerate}

We summarize the detailed procedure in Algorithm \ref{alg:online_RBMLE}.

\begin{algorithm}[t]
    \DontPrintSemicolon
       \textbf{initialization:}  $\pi^{(0)}$.  \\
       \For{$t=0,1,2,\cdots$}
        {
            Sample  $x^{(t)}\sim \rho$, $y_1^{(t)}, y_2^{(t)} \sim \pi^{(t)}(\cdot|x^{(t)})$. \\
            Obtain the preference between $(x^{(t)}, y_1^{(t)})$ and $(x^{(t)}, y_2^{(t)})$ from some oracle. Denote the comparison outcome by $(x^{(t)}, y_+^{(t)}, y_-^{(t)})$.\\
            Update policy $\pi$ as
            \begin{align}   \label{eq:online_RBMLE}
                \pi^{(t+1)} &= \argmin_{\pi  } \Big\{-\sum_{s=1}^{t}{\log \sigma\Big(\beta\log\frac{\pi(y_+^{(s)}|x^{(s)})}{\pi_{\text{ref}}(y_+^{(s)}|x^{(s)})} - \beta\log\frac{\pi(y_-^{(s)}|x^{(s)})}{\pi_{\text{ref}}(y_-^{(s)}|x^{(s)})}\Big)} \nonumber \\
                &\qquad \qquad\qquad + \alpha \beta  \exlim{x \sim \rho, y \sim \dnorm(\cdot|x)}{\log \pi(y|x) - \log\pi_{\text{ref}}(y|x)}\Big\}.
            \end{align}
        }
        \caption{\algabb for online RLHF}
        \label{alg:online_RBMLE}
    \end{algorithm}

\paragraph{Theoretical analysis.} Encouragingly, \algabb admits appealing theoretical guarantees under function approximation.   
For simplicity, we restrict attention to linear approximation of the reward model.
 
\begin{assumption}[Linear Reward]
\label{assump:linear}
We parameterize the reward model by 
\begin{equation}
    r_\theta(x, y) = \innprod{\phi(x, y), \theta},\quad \forall (x, y)\in\mathcal{X}\times \mathcal{Y},
    \label{eq:linear_FA}
\end{equation} where $\phi: \mathcal{X}\times\mathcal{Y} \to \mathbb{R}^d$ is a fixed feature mapping and $\theta \in \mathbb{R}^d$ is the parameters. We assume that $\|\phi(x, y)\|_2\le 1$ for all $(x, y) \in \mathcal{X}\times\mathcal{Y}$, and that $r^\star(x, y)= \innprod{\phi(x, y), \theta^\star}$ for some $\theta^\star$.
\end{assumption}
Under Assumption~\ref{assump:zero_mean} and \ref{assump:linear}, it is sufficient to focus on $\theta\in\Theta$ where
\begin{equation}
    \Theta = \Big\{   \theta  \in \mathbb{R}^d:  \, \exlim{\substack{x\sim\rho, y \sim \dnorm(\cdot| x)}}{\innprod{\phi(x, y), \theta}} = 0 \Big\}.
    \label{eq:THETA}
\end{equation}

The next theorem demonstrates that Algorithm \ref{alg:online_RBMLE} achieves $\widetilde{\mathcal{O}}(\sqrt{T})$ cumulative regret under mild assumptions. The proof is provided in Appendix~\ref{sec:pf_thm_online}. The proof logic follows from
that of \citep{liu2024maximize}.
\begin{theorem}\label{thm_online}
Under Assumptions \ref{assump:zero_mean} and \ref{assump:linear}, let $r_{\theta^{(t)}}\in \Theta$ denote the corresponding reward model for $\pi^{(t)}$.    Assume that $\|\theta^\star\|_2 \le C$ and $\|\theta^{(t)}\|_2 \le C, \forall t \ge 0$ for some $C > 0$.
     Then with probability $1-\delta$ we have 
    \begin{align*}
        \mathsf{Regret} &\coloneqq \sum_{t=1}^{T} \big[J^\star(r^\star) - J(r^\star, \pi^{(t)}) \big] \le \widetilde{\mathcal{O}}(\exp(2C+C/\beta)\sqrt{\kappa d T}),
    \end{align*}
    with $\alpha = \dfrac{1}{\exp(2C+C/\beta)}\sqrt{\dfrac{T}{\kappa\min\{d\log T, T\}}}$ and $\kappa = \sup_{x,y}\dfrac{\dnorm(y|x)}{\pi_{\text{ref}}(y|x)}$.
\end{theorem}
Theorem~\ref{thm_online} shows that \algabb achieves the same $\widetilde{O}(\sqrt{T})$ regret for online RLHF as its counterparts in standard contextual bandits with scalar rewards and using UCB for exploration \citep{lattimore2020bandit}.

\begin{remark}
    The analysis naturally extends to allowing mini-batch samples of size $M$ in every iteration, yielding an improved regret bound scaled by $1/
    \sqrt{M}$ and $\alpha$ scaled by $\sqrt{M}$.
\end{remark}

\subsection{Offline RLHF: algorithm and theory}
In offline RLHF,  a fixed offline preference dataset is collected $\mathcal{D} \coloneqq \{x^{i}, y_+^{i}, y_-^{i}\}_{i=1}^{N}$, where $x^i\sim\rho$, $y^i\sim \pi_{\mathsf{b}}(\cdot|x)$ are sampled from a behavior policy $\pi_{\mathsf{b}}$, such as $\pi_{\text{sft}}$ from SFT. The proposed \algabb for offline RLHF consists of one pass through the reward and policy learning phases, i.e.,
\begin{equation} \label{eq:RBMLE_offline}
    \widehat{r} = 
    \arg\min_{r\in\mathcal{R}} \; \{\ell(r, \mathcal{D}) + \alpha \cdot J^\star(r)\}
\qquad   \mbox{and} \qquad
    \widehat{\pi} = \arg\max_{\pi} J(\widehat{r}, \pi),
\end{equation}
which discourages over-optimization of the reward function given the limited offline preference data. 
In the same vein as deriving \eqref{eq:online_RBMLE}, and by leveraging \eqref{eq:Jstar}, we obtain the direct policy update rule:
\begin{align} 
\widehat{\pi} & = \arg\min_{\pi} \Big\{-\sum_{i=1}^{N}{\log \sigma\Big(\beta\log\frac{\pi(y_+^{i}|x^{i})}{\pi_{\text{ref}}(y_+^{i}|x^{i})} - \beta\log\frac{\pi(y_-^{i}|x^{i})}{\pi_{\text{ref}}(y_-^{i}|x^{i})}\Big)} \notag \\
            &\qquad\qquad\qquad - \alpha \beta  \exlim{x \sim \rho, y \sim \dnorm(\cdot|x)}{\log \pi(y|x) - \log\pi_{\text{ref}}(y|x)}\Big\}.
\end{align}
We summarize the detailed procedure in Algorithm \ref{alg:offline_RBMLE}. When $\dnorm$ is set to $\pi_{\text{ref}}$, the regularization term becomes the KL divergence between $\pi$ and $\pi_{\text{ref}}$, which is reminiscent of a popular choice in offline RL practice  \citep{kumar2020conservative}. Another heuristic choice is to set $\dnorm$ to the marginalized positive answer distribution from the dataset, \ie, $(x, y_+)\sim\mathcal{D}$, which leads to a similar objective in \citep{pal2024smaug}.

\begin{algorithm}[t]
    \DontPrintSemicolon
   \textbf{input:}  offline preference data $\mathcal{D}$ of size $N$.  \\
    Get policy $\widehat{\pi}$ by optimizing
    \begin{align*}
        \widehat{\pi} &= \arg\min_{\pi} \Big\{-\sum_{i=1}^{N}{\log \sigma\Big(\beta\log\frac{\pi(y_+^{i}|x^{i})}{\pi_{\text{ref}}(y_+^{i}|x^{i})} - \beta\log\frac{\pi(y_-^{i}|x^{i})}{\pi_{\text{ref}}(y_-^{i}|x^{i})}\Big)}\\
        &\qquad\qquad\qquad - \alpha \beta  \exlim{x \sim \rho, y \sim \dnorm(\cdot|x)}{\log \pi(y|x) - \log\pi_{\text{ref}}(y|x)}\Big\}.
    \end{align*}
    \caption{\algabb for offline RLHF} \label{alg:offline_RBMLE}
\end{algorithm}

\paragraph{Saddle-point characterization and pessimism.} We first illustrate that \algabb indeed executes the principle of pessimism in a complementary manner to the standard approach of pessimism, which finds a policy that maximizes the worst-case value function over a confidence set. In particular, this strategy 
\citep{uehara2021pessimistic} obtains a policy by solving
\begin{equation}
\widehat{\pi}_{\mathsf{LCB}} = \arg\max_{\pi}\min_{r\in\mathcal{R}_\delta} J(r, \pi)
    \label{eq:authentic_pessimism}
\end{equation}
where the confidence set $\mathcal{R}_\delta$ is typically set to $\{r:\ell(r, \mathcal{D}) \le \ell(r_{\mathsf{MLE}}, \mathcal{D}) + \delta\}$ or $\{r: \mathsf{dist}(r, r_{\mathsf{MLE}}) \le \delta\}$ for some $\delta > 0$ and s distance measure $\mathsf{dist}$.  
Turning to \algabb, note that by \eqref{eq:RBMLE_offline} we have 
\begin{align}
    \widehat{r} &= \arg\min_{r} \big\{\ell({r}, \mathcal{D}) + \alpha J^\star({r}) \big\} = \arg\min_{r} \max_{\pi} \big\{\ell({r}, \mathcal{D}) + \alpha J({r}, \pi)\big\}.
    \label{eq:pessimism_r}
\end{align}
Since $\ell({r}, \mathcal{D}) + \alpha J({r}, \pi)$ is strongly concave over $\pi$, and convex over $r$, it allows us to formulate $(\widehat{r},\widehat{\pi})$ as a saddle point in the following lemma. The proof is given in Appendix~\ref{sec:pf_saddle_point}.
\begin{lemma}
    $(\widehat{r},\widehat{\pi})$ is a saddle point of the  objective $\ell({r}, \mathcal{D}) + \alpha J({r}, \pi)$, \ie, for any $(r', \pi')$, we have
    \begin{equation*}
    \begin{cases}
        \ell(\widehat{r}, \mathcal{D}) + \alpha J(\widehat{r}, \widehat{\pi}) \le \ell({r}', \mathcal{D}) + \alpha J({r}', \widehat{\pi})\\
        \ell(\widehat{r}, \mathcal{D}) + \alpha J(\widehat{r}, \widehat{\pi}) \ge \ell(\widehat{r}, \mathcal{D}) + \alpha J(\widehat{r}, \pi')
    \end{cases}.
    \end{equation*}
    \label{lemma:saddle_point}
\end{lemma}
As such, the policy obtained by \algabb can be equivalently written as
\begin{align}
    \widehat{\pi} &\in \arg\max_{\pi} \min_{r} \Big\{J({r}, \pi) + \frac{1}{\alpha}\ell({r}, \mathcal{D})\Big\} = \arg\max_{\pi} \min_{r\in \mathcal{R}_{\delta(\pi, \alpha)}} J({r}, \pi),
    \label{eq:counterfeit_pessimism}
\end{align} 
where $\mathcal{R}_{\delta(\pi, \alpha)}$ is the constraint set $\{r:\ell(r, \mathcal{D}) \le \ell(r_{\mathsf{MLE}}, \mathcal{D}) + \delta(\pi, \alpha)\}$ such that the constrained optimization problem $\min_{r\in \mathcal{R}_{\delta(\pi, \alpha)}} J({r}, \pi)$ is equivalent to the regularized problem $\min_{r} \big\{J({r}, \pi) + \frac{1}{\alpha}\ell({r}, \mathcal{D})\big\}$. In view of the similarity between the formulations \eqref{eq:authentic_pessimism} and \eqref{eq:counterfeit_pessimism}, we conclude that \algabb implements the  pessimism principle \eqref{eq:authentic_pessimism} in an oblivious manner without explicitly estimating the uncertainty level,  justifying popular practice as a valid approach to pessimism \citep{kumar2020conservative}.

\paragraph{Theoretical analysis.} 
The next theorem establishes the sub-optimality gap of \algabb with linear function approximation under mild assumptions.
The proof is given in Appendix~\ref{sec:pf_thm_offline}.
\begin{theorem}
Under Assumptions \ref{assump:zero_mean} and \ref{assump:linear}, let $\widehat{\theta}\in \Theta$ denote the corresponding reward model for $\widehat{\pi}$.
Assume that $\|\theta^\star\|_2 \le C$ and $\|\widehat{\theta}\|_2 \le C$ for some $C > 0$. Let $\alpha = \sqrt{N}$ and $\delta\in (0,1)$. With probability $1-\delta$, we have
    \begin{align*}
    J^\star(r^\star) - J(r^\star, \widehat{\pi}) 
    &\le \mathcal{O}\Bigg(\frac{C_1}{\sqrt{N}}\cdot\Big\|\exlim{\substack{x \sim \rho,\\y\sim\pi^\star(\cdot|x)}}{\phi(x, y)}\Big\|_{(\Sigma_\mathcal{D} + \lambda I)^{-1}} + \frac{C_2}{\sqrt{N}}\Bigg),
\end{align*}
where $
    \Sigma_\mathcal{D} = \frac{1}{N}\sum_{i=1}^{N}(\phi(x^{i},y_+^i)-\phi(x^{i},y_-^i))(\phi(x^{i},y_+^i)-\phi(x^{i},y_-^i))^\top$ is the feature sample  covariance matrix, 
$\lambda = 1/N$, $C_1 = \exp(C)\Big(\sqrt{{{d + \log (1/\delta)}}} + \kappa_{\mathcal{D}}\Big)+ C $
and $C_2 =  {\exp(C)}\kappa_{\mathcal{D}}^2 + C\kappa_{\mathcal{D}}  + 1$. 
Here,
\begin{equation*}
    \kappa_{\mathcal{D}} = \Big\|\exlim{\substack{x\sim\rho,\\ y \sim \widehat{\pi}(\cdot| x)}}{\phi(x, y)} - \exlim{\substack{x\sim\rho,\\ y \sim \dnorm(\cdot| x)}}{\phi(x, y)}\Big\|_{(\Sigma_\mathcal{D} + \lambda I)^{-1}} \le 4 (\lambda_{\min}(\Sigma_\mathcal{D}) + \lambda)^{-1}.
\end{equation*}
\label{thm:offline}
\end{theorem}

Theorem~\ref{thm:offline} establishes that \algabb achieves the same rate of $\widetilde{\mathcal{O}}(1/\sqrt{N})$ as standard offline RL, as long as the offline dataset $\mathcal{D}$ has sufficient coverage. 
We remark that $\Big\|\exlim{\substack{x \sim \rho,\\y\sim\pi^\star(\cdot|x)}}{\phi(x, y)}\Big\|_{(\Sigma_\mathcal{D} + \lambda I)^{-1}}$ is reminiscent of the standard single-policy concentratability coefficient in  offline RL, which measures the distribution shift between the offline dataset and the optimal policy \citep{zhu2023principled}.

\section{Token-level \algabb}\label{sec:tokenMDP}
  
Recently, \citet{rafailov2024r,zhong2024dpo} offered an interpretation of DPO using the token-level Markov Decision Process (MDP), aiming at reconciling the gap between the practical fine-tuning of LLMs at the token level and the theoretical formulation of DPO at the sentence level. Fortunately, this interpretation does not require algorithmic modifications. Below we provide a short discussion and highlight that \algabb can be similarly interpreted using the token-level MDP following the setups in \citet{rafailov2024r,zhong2024dpo}.

\paragraph{Token-level MDP and preference modeling.}  Recall that in LLMs, the prompt $x$ can be broken into a sequence of tokens, e.g., $x=(x_0,\ldots,x_m)$, from a fixed discrete vocabulary $\cA$. 
We define the token-level MDP as a tuple $\cM = (\cS, \cA, P, r^{\star}, H)$, where $H$ is the horizon length, i.e., the longest possible number of tokens in a sentence. 
The state space $\cS$ consists of all the possible token combinations of length $H$, and the transition kernel is deterministically defined as follows. 
\begin{enumerate}
\item The initial state is defined by the prompt $x$ as $s_0=\{x_0, \cdots, x_m\}$; 
\item Given the response $y=\{ y_0, \cdots , y_{i-1}\}$ up to the $i$-th token, the state at step $i$ is defined as  $s_i = \{x_0, \cdots, x_m, y_0, \cdots , y_{i-1}\}$;
\item Upon an action of the LLM for the next token $a_i = y_i$, the next state at the token-level MDP deterministically becomes $s_{i+1} = (s_i, a_i) = (x_0, \cdots, x_m, y_0, \cdots , y_{i})$.
\end{enumerate}
We assume that the last token of a sentence, the EOS token, is absorbing, such that the token-level MDP stays in the corresponding state as soon as the last action is the EOS token.  With slight abuse of notation from earlier sections, the reward function $r^{\star}(s,a)$ defines the ground truth reward at state $s$ upon action $a$.

Given a pair of trajectories $\tau_1=\{s_0,a_0^1,\ldots,s_{H-1}^1,a_{H-1}^1,s_H^1\}$  and $\tau_2=\{s_0,a_0^2,\ldots,s_{H-1}^2,a_{H-1}^2,s_H^2\}$, the corresponding Bradley-Terry preference model~\citep{bradley1952rank} is
\begin{align*} 
    \mathbb{P}(\tau_1 \succ \tau_2) &= \frac{\exp\left(\sum_{i=0}^{H-1}   \rgt(s_i^1, a_i^1)\right)}{\exp\left(\sum_{i=0}^{H-1}  r^{\star}(s_i^1, a_i^1)\right)+\exp\left(\sum_{i=0}^{H-1}  r^{\star}(s_i^2, a_i^2)\right)}\notag\\
    &=\sigma\left(\sum_{i=0}^{H-1}   r^{\star}(s_i^1, a_i^1)-\sum_{i=0}^{H-1}  r^{\star}(s_i^2, a_i^2)\right),
\end{align*}
A preference data sample is denoted by a tuple $(x, \tau_+, \tau_-)$, where $\tau_+$ (resp.  $\tau_{-}$) is the preferred (resp. unpreferred) answer in the comparison. Given a preference dataset $\mathcal{D}  $ composed of independent samples, 
The negative log-likelihood can be defined as
\begin{equation}\label{eq:nll}
    \ell(r,\cD)\coloneqq-\sum_{(\tau_{+},\tau_{-})\in\cD}\log \sigma\left(\sum_{i=0}^{H-1}  r(s_i^{+}, a_i^{+})-\sum_{i=0}^{H-1}  r(s_i^{-}, a_i^{-})\right),
\end{equation}
where $(s_i^{+},a_i^{+})$ (resp. $(s_i^{-},a_i^{-})$) are the state-action pairs in the trajectory $\tau_{+}$ (resp. $\tau_{-}$).

\paragraph{Token-level RL fine-tuning.} Let the entropy of policy $\pi$ under initial state distribution $s_0 \sim \rho$ be defined as
$$ \cH(\rho,\pi)\coloneqq -  \underset{s_0\sim\rho,\atop a_i\sim\pi(\cdot|s_i) }{\EE} \left[\sum_{i=0}^{H-1}  \log\pi(a_i|s_i)\right], $$
and  $\cH(s,\pi)$ be the entropy when the initial state $s_0 =s$.
Given a reward function $r$, we define the KL-constrained RL objective against the reference policy $\piref$ as \citep{rafailov2024r}:
\begin{align}\label{eq:KL_constrained_RL_obj}
   \pi_r\coloneqq \argmax_{\pi}   J(r,\pi)&\coloneqq\underset{s_0\sim\rho,\atop a_i\sim\pi(\cdot|s_i)}{\EE}\Big[\sum_{i=0}^{H-1}  ( \underbrace{ r(s_i,a_i)+\beta\log \piref(a_i|s_i)}_{r_{\beta}(s_i, a_i)})\Big] +\beta\cH(\rho,\pi) ,
\end{align}
where we denote $r_\beta:\cS\times\cA\rightarrow \mathbb{R}$ as follows:
\begin{equation}\label{eq:R}
    \forall (s,a)\in\cS\times\cA:\quad r_\beta(s,a)\coloneqq r(s,a)+\beta\log\piref(a|s),
\end{equation}
which can be seen as the actual token-wise reward function optimized by the LLM. 
 
\paragraph{Token-level DPO.} The KL-constrained RL objective~\eqref{eq:KL_constrained_RL_obj} has a closed-form solution~\citep{nachum2017bridging,cen2022fast} given by
\begin{equation}\label{eq:opt_pi}
    \forall (s,a)\in\cS\times\cA:\quad
    \pi_r(a|s)=\exp((Q_{\beta}^\star(s,a)-V_{\beta}^\star(s))/\beta),
\end{equation}
where $V_{\beta}^\star:\cS\rightarrow \mathbb{R}$ and $Q_{\beta}^\star: \cS\times \cA\rightarrow \mathbb{R}$ are the optimal soft value and Q functions, respectively,
\begin{equation}\label{eq:soft_V*}
    \forall s\in\cS:\quad V_{\beta}^\star(s)\coloneqq\underset{a_i\sim\pi_r(\cdot|s_i) }{\EE}\left[\sum_{i=0}^{H-1} r_\beta(s_i,a_i)|s_0 = s\right]  + \beta \cH(s,\pi_r)
\end{equation}
denote the optimal soft value function w.r.t. the reward function $r_\beta$, and  
\begin{equation}\label{eq:soft_Q*}
    \forall (s,a)\in\cS\times\cA:\quad Q_{\beta}^\star(s,a)\coloneqq r_\beta(s,a) +  V_{\beta}^\star (s'),
\end{equation}
where $s' = (s,a)$ is the deterministic next state. Plugging \eqref{eq:soft_Q*} and \eqref{eq:R} into \eqref{eq:opt_pi} implies that, for any trajectory $\tau=\{s_0,a_0,\cdots,a_{H-1},s_H\}$, \citet{rafailov2024r} shows
\begin{align}\label{eq:sum_r}
    \sum_{i=0}^{H-1} r(s_i,a_i)& = \sum_{i=0}^{H-1}  \left(Q_{\beta}^\star(s_i,a_i)-\beta\log\piref(a_i|s_i)-  V_{\beta}^\star(s_{i+1})\right)\notag\\
     &=Q_{\beta}^\star(s_0,a_0)-\beta\log\piref(a_0|s_0)+\sum_{i=1}^{H-1}  \left(Q_{\beta}^\star(s_i,a_i)-V_{\beta}^\star(s_i)-\beta\log\piref(a_i|s_i)\right)\notag\\
  & = V_\beta^{\star}(s_0)+ \beta\sum_{i=0}^{H-1}  \log \frac{\pi_r(a_i|s_i)}{\piref(a_i|s_i)},
\end{align}
where the second line uses $V_{\beta}^{\star}(s_H)=0$ at the terminal state.
 Thus the DPO loss (which is the negative log-likelihood loss) could be written as
\begin{equation}\label{eq:DPO_loss_token}
    \cL(\pi,\cD)=-
    \sum_{(\tau_+,\tau_-)\in\cD}\log \sigma\left(\beta\sum_{i=0}^{H-1}  \log\frac{\pi(a_i^+|s_i^+)}{\piref(a_i^+|s_i^+)}-\beta\sum_{i=0}^{H-1}  \log\frac{\pi(a_i^-|s_i^-)}{\piref(a_i^-|s_i^-)}\right).
\end{equation}

\paragraph{Token-level \algabb.} With slight abuse of notation, define
\begin{equation}\label{eq:pi_r_and_J_star}
 J^\star(r)\coloneqq\max_\pi J(r,\pi),
\end{equation}
which is used as the bias term in regularizing the reward estimation in \algabb. Again, we impose the following assumption to deal with the shift ambiguity issue caused by the Bradley-Terry model:
\begin{assumption}
    \label{asmp:zero_mean_token}
We assume that $r^\star \in\mathcal{R}$, where  
\begin{equation}
    \mathcal{R} = \bigg\{r:
    \underset{s_0\sim\rho,\atop a_i\sim\dnorm(\cdot|s_i) }{\EE}
    \sum_{i=0}^{H-1}  r(s_i,a_i) = 0
    \bigg\}.
\end{equation}
Here, $\rho$ is the prompt distribution and $\dnorm$ is a fixed calibration distribution independent of the algorithm.
\end{assumption}

Combining \eqref{eq:KL_constrained_RL_obj} with \eqref{eq:soft_V*}, similar to previous derivations, we have
\begin{align}\label{eq:J*r}
J^\star(r)
    &=\underset{s_0\sim\rho }{\EE}\left[V^\star_{\beta}(s_0)\right]\notag\\
  &=  \underset{s_0\sim\rho,\atop a_i\sim\dnorm(\cdot|s_i) }{\EE}\left[V^\star_{\beta}(s_0)\right]\notag\\
    &=\underset{s_0\sim\rho,\atop a_i\sim\dnorm(\cdot|s_i) }{\EE}
    \left[\sum_{i=0}^{H-1}  r(s_i,a_i)-\beta\sum_{i=0}^{H-1}   \log\frac{\pi_r(a_i|s_i)}{\piref(a_i|s_i)}\right]\notag\\
    & = -\beta\underset{s_0\sim\rho,\atop a_i\sim\dnorm(\cdot|s_i) }{\EE}\left[\sum_{i=0}^{H-1}  \log\frac{\pi_r(a_i|s_i)}{\piref(a_i|s_i)}\right],
\end{align}
where the penultimate line uses \eqref{eq:sum_r}, and the last line uses Assumption~\ref{asmp:zero_mean_token}.

Consequently, the token-level \algabb can be rewritten as 
\begin{align}\label{eq:pi_token_vpo}
       \pi_{\mathsf{VPO}}      &=\arg\min_{\pi} \Bigg\{
    -
    \sum_{(\tau_+,\tau_-)\in\cD}\log \sigma\left(\beta\sum_{i=0}^{H-1}   \log\frac{\pi(a_i^+|s_i^+)}{\piref(a_i^-|s_i^-)}-\beta\sum_{i=0}^{H-1} \log\frac{\pi(a_i^- | s_i^-)}{\piref(a_i^-|s_i^-)}\right)\notag\\
    &\qquad\qquad\qquad + \mathsf{sign} \cdot\alpha \beta  \underset{s_0\sim\rho,\atop a_i\sim\dnorm(\cdot|s_i) }{\EE}\left[\sum_{i=0}^{H-1}  \log\frac{\pi(a_i|s_i)}{\piref(a_i|s_i)}\right]\Bigg\}.
\end{align}

\section{Experiments}
 
In this section, we evaluate the proposed~\algabb on both synthetic bandits and RLHF for LLMs, in online and offline settings. 

\subsection{Synthetic bandit problems} 
 
We evaluate the proposed methods on two synthetic problems: i) a multi-armed bandit problem, and ii) a linear contextual bandit problem.
\paragraph{Multi-armed bandit (MAB) problem.} In this scenario, we set $|\mathcal{X}| = 1$ and $|\mathcal{Y}| = 10$.  For each $y\in\mathcal{Y}$, the ground truth reward $r^\star(x, y)$ is randomly generated i.i.d. from a uniform distribution $U([0, 1])$. The policy is parameterized by $\pi_{\theta}(\cdot|x) = \mathsf{softmax}(\theta(x,\cdot))$, where $\theta\in\mathbb{R}^{10}$. The reference policy $\pi_{\text{ref}}$ is set to $\pi_{\theta_{\text{ref}}}$ with $\theta_{\text{ref}}(x, y)$ sampled i.i.d. from $U([0, 1])$. 
\paragraph{Linear contextual bandit problem.} Here, we set $\mathcal{X} = \mathbb{R}^2$, $|\mathcal{Y}| = 50$. For each $(x, y)$ pair, the ground truth reward is given by $r^\star(x, y) = \left\langle\phi(x, y), \theta^\star\right\rangle$, where $\theta^\star\in\mathbb{R}^{10}$ is randomly sampled from $U([0, 1])$, and the feature vector $\phi(x, y)$ is the output of the hidden layer of a fixed two-layer MLP, with the input given by the concatenation of $x$ and the one-hot encoding of $y$. The activation function is set to tanh. The context vector $x$ is drawn from standard normal distribution. We focus on log-linear policy class $\pi_\theta(\cdot|x) = \mathsf{softmax}(\langle\theta, \phi(x, \cdot)\rangle)$, and set $\pi_{\text{ref}} = \pi_{\theta_{\text{ref}}}$ with $\theta_{\text{ref}}(x, y)$ sampled i.i.d. from $U([0, 1])$. 

For both problem we set $\pi_{\text{ref}} = \pi_{\mathsf{b}}= \dnorm$ and use mini-batch sample of size 5 in every iteration. We approximately solve the optimization problems by performing $20$ AdamW optimization steps with learning rate $0.01$ and weight decay rate $0.01$ in every iteration for the online setting and $1000$ steps for the offline setting. We set $\beta=5$ for the online linear contextual bandit problem and $\beta=1$ for all other experiments to better illustrate the performance differences.

\begin{figure*}[ht]
    \centering
    \begin{tabular}{cc}
   \includegraphics[width=0.46\textwidth]{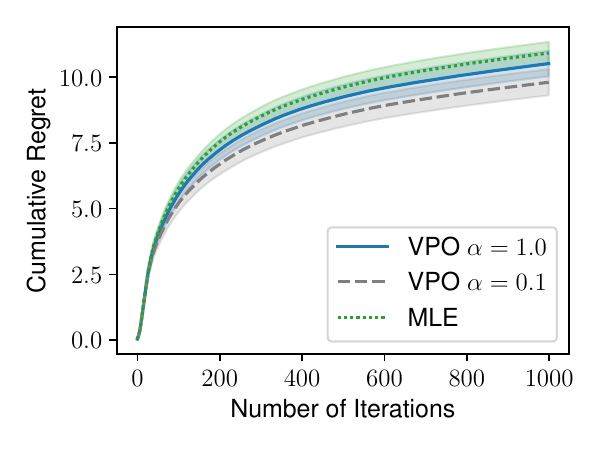} &  \includegraphics[width=0.46\textwidth]{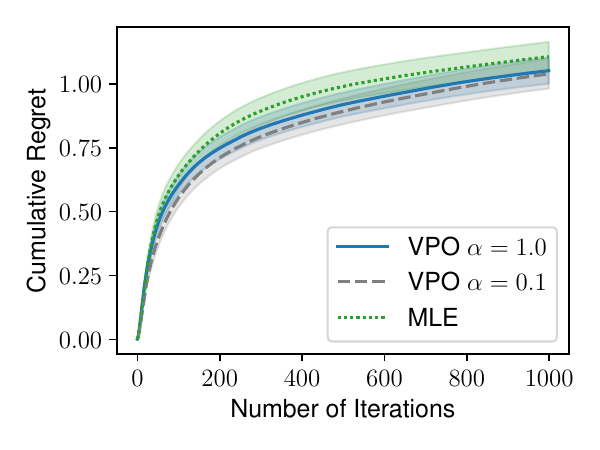} 
    \end{tabular}
    \caption{The cumulative regret v.s. number of iterations plot of \algabb and MLE-MAB methods in the online MAB problem (left panel) and online linear contextual bandit problem (right panel), respectively. 
    }

    \label{fig:synthetic_online}
\end{figure*}

\begin{figure*}[ht]
    \centering
    \begin{tabular}{cc}
   \includegraphics[width=0.46\textwidth]{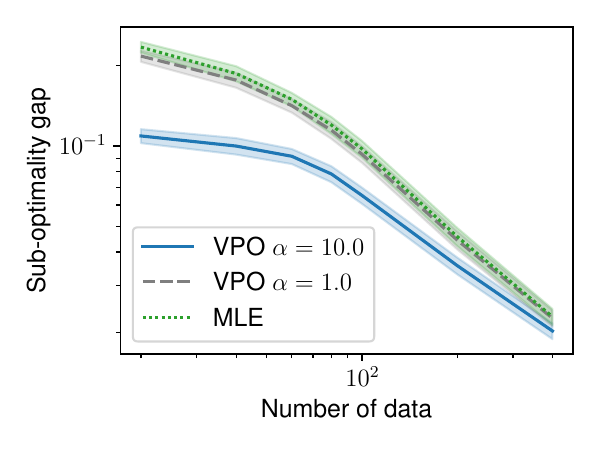} &  \includegraphics[width=0.46\textwidth]{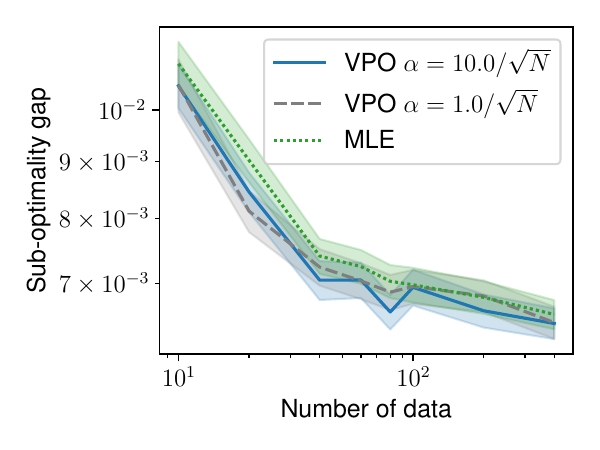} 
    \end{tabular}
    \caption{The sub-optimality gap v.s. number of data plot of \algabb and MLE-MAB methods in the offline MAB problem (left panel) and offline linear contextual bandit problem (right panel), respectively. 
    }

    \label{fig:synthetic_offline}
\end{figure*}

We plot the average results over 50 independent runs for both experiments in Figure~\ref{fig:synthetic_online} and Figure~\ref{fig:synthetic_offline}. As demonstrated in Figure~\ref{fig:synthetic_online}, an appropriate choice of $\alpha$ allows our method to outperform the model-based MAB with MLE baseline in the long-term performance of cumulative regret, at the cost of slightly increased cumulative regret in the first 100 iterations. This highlights the effectiveness of the \algabb in achieving more principled exploration-exploitation trade-off. For the offline setting, Figure~\ref{fig:synthetic_offline} demonstrates that the performance of both MLE-MAB and \algabb improves as the number of offline data increases. However, \algabb achieves a consistently lower sub-optimality gap compared with that of MLE-MAB.

\subsection{RLHF for LLMs} 

We further evaluate the pessimistic/optimistic~\algabb for LLMs in offline and online setting, respectively. In both settings, the proposed \algabb demonstrates strong performances over the baselines.

\paragraph{Offline setting.} In this setting, we test pessimistic \algabb on \texttt{ARC-Challenge} task~\citep{clark2018think}, which contains $7787$ multiple-choices questions from multiple science subjects. We evaluate the performances on the \texttt{ARC-Challenge} test set, which contains $1172$ questions. 
The data set only provides ground truth answer for each question. To construct the preference pairs and their labels, for each correct response in the training split, we create three pairs of comparison between the correct answer and each incorrect answer.

We emphasize that our goal is to evaluate the RLHF algorithm designs for LLMs, rather than pushing LLM towards state-of-the-art performance. To demonstrate the advantages of the proposed~\algabb, we conduct comparison with several offline RLHF baselines (DPO~\citep{rafailov2023direct} and IPO~\citep{azar2024general}) on  several LLMs, including \textsc{Llama2-7b-chat}, \textsc{Llama2-13b-chat}~\citep{touvron2023llama} and \textsc{Flan-T5-xl}~\citep{chung2022h}. 
For fair comparison, we keep all the experiment settings and prompts the same for every RLHF algorithm. We did not apply any additional chain-of-thought reasoning to avoid compounding factors affecting the RLHF performances. 
We tuned the hyperparameters for both the proposed~\algabb and the baselines on the validation set to achieve their best performances. For detailed hyperparameters setup, please refer to~Appendix~\ref{appendix:exp_details}.

\begin{figure*}[ht]
    \centering
 
    \begin{tabular}{ccc}
   \includegraphics[width=0.31\textwidth, trim = {10 0 10 0}, clip]{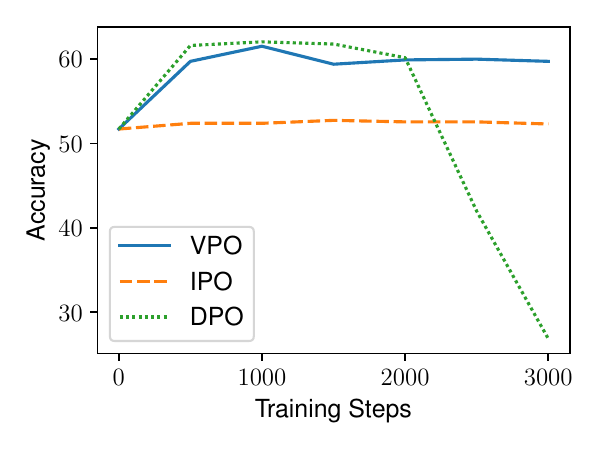} &     \includegraphics[width=0.31\textwidth, trim = {10 0 10 0}, clip]{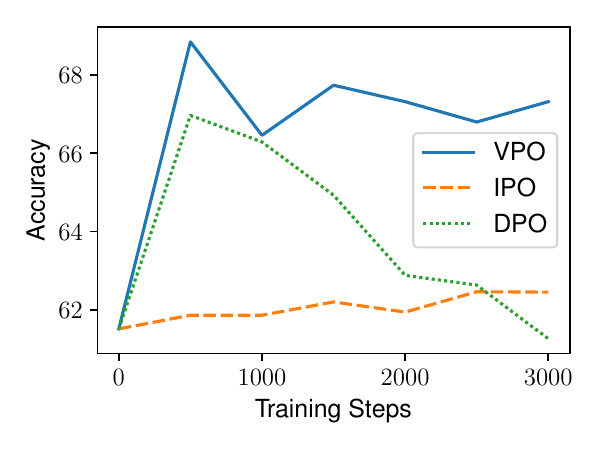} &  \includegraphics[width=0.31\textwidth, trim = {10 0 10 0}, clip]{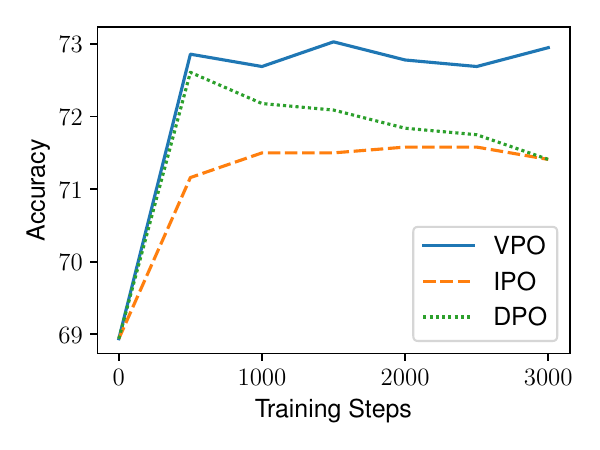}  \\
   (a) \textsc{Llama2-7b-chat} & (b) \textsc{Llama2-13b-chat} & (c) \textsc{Flan-T5-xl}
    \end{tabular}
 
    \caption{The accuracy of the \textsc{Llama2-7b-chat}, \textsc{Llama2-13b-chat} and \textsc{Flan-T5-xl} policies trained by \algabb and other baselines (DPO and IPO) on \texttt{ARC-challenge}, respectively. The proposed pessimistic~\algabb performs consistently strong, and avoids over-optimization.  
    }
    \label{fig:offline}
\end{figure*}

The performances are illustrated in~\figref{fig:offline}.
As we can see, the proposed~\algabb method demonstrates significantly better performance over the existing baselines on the three models, verifying the benefits across different models. In particular, the performance benefit becomes more evident for larger models.
Another important observation is that the proposed~\algabb method is more robust to over-optimization~\citep{gao2023scaling}. In the experiment, the performances of DPO significantly drops after $1000$ iterations, and the longer DPO is trained, the worse it performs. In contrast, \algabb consistently maintains the performances, avoiding the overoptimization issue and justifying the implicit robustness of pessimism as we revealed in~\eqref{eq:pessimism_r}.   
 
\paragraph{Online setting.} For online setting, we conduct two distinct experimental setups. The first, referred to as {\bf Buffer}, adopts the experimental setup in Online AI Feedback (OAIF) ~\citep{guo2024direct}, with an additional buffer used to sample the data for the exploration part of the VPO loss. 
The second setup, referred to as {\bf Iterative}, adopts the experimental setup in ~\citep{zhang2024selfexploring}, relying on an online iterative training framework.

\paragraph{Buffer.}
In these set of experiments, we adopt OAIF experimental setup ~\citep{guo2024direct} where the preference data is gathered by online sampling from the policy and labeled through online feedback.
We also introduce a buffer that stores the labeled preferences and is used to sample the data for the exploration term in the VPO loss.
We adopt \textsc{PaLM2-XXS} language model ~\citep{anil2023palm} as policy, initialized by supervised finetuning, denoted as SFT model.
We exploit another \textsc{PaLM2-XS} model as the LLM annotator to provide online feedback.
We evaluate the performance of optimistic~\algabb and compare its performance to Online DPO ~\citep{guo2024direct}.
We choose \texttt{TL;DR} task~\citep{stiennon2020learning} and extract its prompts for the input of preference data.
Similar to ~\citep{guo2024direct}, we use \emph{Detailed 0-shot} prompt from~\cite{lee2023rlaif}. 
The prompts we used and how we get preference scores are detailed in Appendix~\ref{appendix:exp_details}.
We emphasize our algorithm is agnostic to human or AI feedback. 

\begin{figure*}[ht]
    \centering
 
      \begin{subfigure}[t]{.4\linewidth}
    \centering\includegraphics[width=\linewidth]{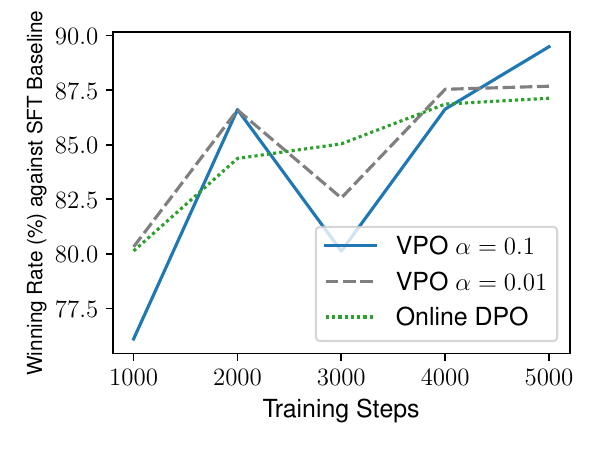}
 
    \newsubcap{Win rate of~\algabb and Online DPO against the SFT baseline on \texttt{TL;DR} task. }
        \label{fig:online_curve}
  \end{subfigure}\quad
  \begin{subfigure}[t]{.4\linewidth}
    \centering\includegraphics[width=\linewidth]{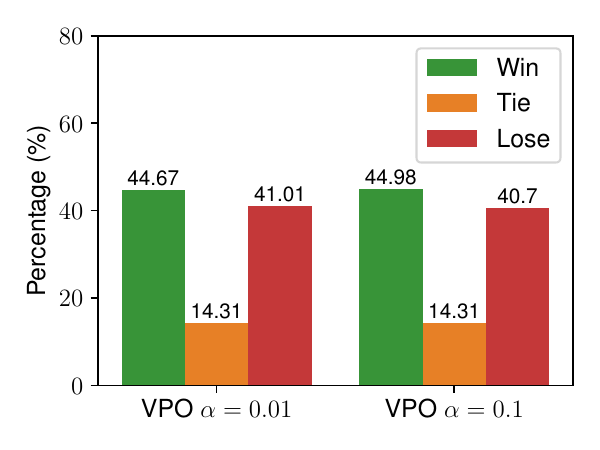}
 
    \newsubcap{Win/tie/loss rate of \algabb with different exploration rate $\alpha = \{0.01, 0.1\}$, directly against Online DPO. }
            \label{fig:online_bar}
  \end{subfigure}
\end{figure*}

As a sanity check, we track the win rate of~\algabb and Online DPO against the SFT baseline on~\texttt{TL;DR} during training in~\figref{fig:online_curve}. For ablation purpose, we vary the exploration weight $\alpha = \{0.01, 0.1\}$ in the optimistic~\algabb. 
One significant observation is that although all the online RLHF algorithms follow the increase trend, the win-rate against SFT of the optimistic~\algabb has larger oscillation, comparing to Online DPO. And the oscillation reduces, with $\alpha$ diminishing. 
Our conjecture is that this behavior is encouraged by the optimistic term in~\algabb, for collecting more unexplored data, which may delay the learning due to the diversity in data. However, as the learning proceeds, the proposed~\algabb outperforms the competitors, because of the coverage of the collected data.  
%
%
To demonstrate the advantages of optimistic \algabb in online setting more directly, we evaluate the win/tie/loss rate against Online DPO head-to-head, as shown in~\figref{fig:online_bar}. This clearly shows that the optimistic~\algabb achieves better performances with larger exploration preference, and thus, consolidates our conclusion that {\em i)}, the simple value-incentivized term makes the exploration practical without uncertainty estimation; and {\em ii)}, exploration is potentially beneficial for better model.

\paragraph{Iterative.} 
We further evaluate the performance of VPO on standard benchmarks AlpacaEval 2.0 \citep{dubois2024alpacafarm, dubois2024length} and MT-Bench \citep{zheng2024judging}, using similar experimental setup in recent literature \citep{zhang2024self, wu2024self}.
We use UltraFeedback\footnote{\url{https://huggingface.co/datasets/HuggingFaceH4/ultrafeedback_binarized}} \citep{cui2023ultrafeedback} as our training dataset which contains around 61k preference pairs of single-turn conversations. 
We split the 61k prompts into four chunks and follow an iterative training approach.
We choose Zephyr-7B-SFT\footnote{\url{https://huggingface.co/HuggingFaceH4/mistral-7b-sft-beta}} as our LLM model. 
We follow the best hyperparameters setup found in \citet{zhang2024self} and compare the results of VPO to DPO reported therein.
We first conduct a single iteration of standard DPO training on the first portion of the training data, referred to as Zephyr-7B-DPO in ~\ref{tab:models}.
We then perform 3 iterations of VPO, each iteration on a different data portion, while using online AI feedback provided by PairRM \citep{jiang2023llm} in between. 
Further details of our experiments are explained in Appendix~\ref{appendix:exp_details}.

The evaluation results are summarized in Table~\ref{tab:models}, filling out the baselines based on \citet{zhang2024self}.
As could been seen, VPO significantly improves the performance of the base model Zephyr-7B-SFT by \textbf{14.52}, achieving the highest length-controlled (LC) Win Rate on the AlpacaEval 2.0 benchmark, beating DPO. This result, 22.53, is competitive to much larger models such as much Yi-34B-Chat, 27.19, and Llama-3-70B-Instruct, 33.17.
Additionally, VPO shows significant improvement on MT-Bench compared to the base model Zephyr-7B-SFT with the increase of \textbf{2.32} while beating DPO.
We further report the results of other iterative post-training algorithms, such as SPIN \citep{chen2024self}, DNO \citep{rosset2024direct}, and SPPO \citep{wu2024self} and show that even though VPO is trained on a weak base model, it achieves comparable results to these baselines. Granular views on a radar chart can be found in Appendix~\ref{appendix:exp_details}.

\begin{center}

\begin{table}[h!]
\centering
\begin{tabular}{c|ccc|ccc} 
\toprule
\multirow{2}{*}{Model} & \multicolumn{3}{c|}{AlpacaEval 2.0} & \multicolumn{3}{c}{MT-Bench} \\  
                                & LC Win Rate & Win Rate & Avg. Len         & Avg & 1st Turn & 2nd Turn  \\ \midrule
Zephyr-7B-SFT & 8.01  & 4.63  & 916  & 5.30 & 5.63 & 4.97 \\ 
Zephyr-7B-DPO & 15.41 & 14.44 & 1752 & 7.31 & 7.55 & 7.07 \\
DPO Iter 1 (Zephyr) & 20.53 & 16.69 & 1598 & 7.53 & 7.81 & 7.25 \\ 
DPO Iter 2 (Zephyr) &22.12 & 19.82 & 1717 & 7.55 & 7.85 & 7.24 \\
DPO Iter 3 (Zephyr) & 22.19 & \textbf{19.88} & 1717 & 7.46 & 7.85 & 7.06 \\
\rowcolor{lightgray}
VPO Iter 1 (Zephyr) & \textbf{22.53} & 19.09 & 1638 & 7.50 & 7.76 & 7.24 \\
\rowcolor{lightgray}
VPO Iter 2 (Zephyr) & 21.84 & 18.78 & 1663 & \textbf{7.62} & 7.93 & \textbf{7.32} \\
\rowcolor{lightgray}
VPO Iter 3 (Zephyr) & 22.15 & 19.58 & 1713 & 7.61 & \textbf{8.01} & 7.21 \\\midrule
SPIN  & 7.23 & 6.54 & 1426 & 6.54 & 6.94 & 6.14 \\
Orca-2.5-SFT & 10.76 & 6.99 & 1174 & 6.88 & 7.72 & 6.02 \\
DNO (Orca-2.5-SFT) & 22.59 & 24.97 & 2228 & 7.48 & 7.62 & 7.35 \\
Mistral-7B-Instruct-v0.2 & 19.39 & 15.75 & 1565 & 7.51 & 7.78 & 7.25 \\
SPPO (Mistral-it) & 28.53 & 31.02 & 2163 & 7.59 & 7.84 & 7.34 \\\midrule
Yi-34B-Chat & 27.19 & 21.23 & 2123 & 7.90 & - & - \\
Llama-3-70B-Instruct & 33.17 & 33.18 & 1919 & 9.01 & 9.21 & 8.80 \\
GPT-4 Turbo (04/09) & 55.02 & 46.12 & 1802 & 9.19 & 9.38 & 9.00 \\\bottomrule
\end{tabular}
\caption{Results on AlpacaEval 2.0 and MT-Bench.}
\label{tab:models}
\end{table}
\end{center}

\section{Conclusion and Discussion}\label{sec:conclusion}
 
In this work, we develop a unified approach to achieving principled optimism and pessimism in online and offline RLHF, which enables a practical computation scheme by incorporating uncertainty estimation implicitly within reward-biased maximum likelihood estimation. Theoretical analysis indicates that the proposed methods mirror the guarantees of their standard RL counterparts, which is furthermore corroborated by numerical results. Important future directions include investigating adaptive rules for selecting $\alpha$ without prior information and more refined analysis on the choice of $\dnorm$.
This work also hints at a general methodology of designing practical algorithms with principled optimism/pessimism under more general RL setups.

\section*{Acknowledgement}
The work of S. Cen, T. Yang and Y. Chi is supported in part by the grants ONR N00014-19-1-2404, NSF DMS-2134080, CCF-2106778 and CNS-2148212. S. Cen is also gratefully supported by Wei Shen and Xuehong Zhang Presidential
Fellowship and JP Morgan AI Research Fellowship.

\bibliographystyle{apalike}
\bibliography{RLHF_refs.bib}

\newpage
 
\appendix

\section{Analysis for the online setting}
\label{sec:pf_thm_online}

\subsection{Proof of Theorem~\ref{thm_online}}
 
For ease of presentation,   
 we assume that $\mathcal{R}$ is finite, i.e., $|\mathcal{R}|<\infty$. The general case can be directly obtained  using a covering number argument, which we refer to 
\citep{liu2024maximize,jin2022power} for interested readers.

We start by decomposing the regret into two parts:
\begin{align}
    \mathsf{Regret} &\coloneqq \sum_{t=1}^{T} \big[J^\star(r^\star) - J(r^\star, \pi^{(t)}) \big]\notag\\
    &= \underbrace{\sum_{t=1}^{T} \big[J^\star(r^\star) - J^\star(r^{(t)}) \big]}_{\textbf{Term (i)}} + \underbrace{\sum_{t=1}^{T} \big[J(r^{(t)}, \pi^{(t)}) - J(r^\star, \pi^{(t)}) \big]}_{\textbf{Term (ii)}} .
    \label{eq:regret_decomp}
\end{align}

\paragraph{Step 1: bounding term (i).}
By the choice of $r^{(t)}$, we have
\begin{align}
    \ell(r^{(t)}, \mathcal{D}^{(t-1)}) - \alpha J^\star(r^{(t)}) \le \ell(r^\star, \mathcal{D}^{(t-1)}) - \alpha J^\star(r^\star).
\end{align}
Rearranging terms, 
\begin{align}
    J^\star(r^\star) - J^\star(r^{(t)}) &\le \frac{1}{\alpha}\big[\ell(r^\star, \mathcal{D}^{(t-1)}) - \ell(r^{(t)}, \mathcal{D}^{(t-1)})\big].
\end{align}
The following lemma is adapted from \cite[Proposition 5.3]{liu2024maximize}, whose proof is deferred to Appendix~\ref{sec:proof_good_wheel}.
\begin{lemma}
Let $\delta \in (0,1)$. With probability $1-\delta$, we have 
\begin{align}
    &\ell(r^\star, \mathcal{D}^{(t-1)}) - \ell(r^{(t)}, \mathcal{D}^{(t-1)}) \notag\\
    &\le -2\sum_{s=1}^{t-1}\exlim{\substack{x \sim \rho,\\ (y_1,y_2)\sim \pi^{(s)}(\cdot| x )}}{D_{\text{H}}^2(\mathbb{P}_{r^{(t)}}(\cdot|x ,y_1 ,y_2 )\,\|\,\mathbb{P}_{r^\star}(\cdot|x ,y_1 ,y_2 ))} + 2 \log(|\mathcal{R}|/\delta).
\end{align} 
Here, $D_{\text{H}}(\cdot \| \cdot)$ is the Hellinger distance, $\mathbb{P}_{r}(\cdot|x,y_1,y_2)$ denotes the Bernoulli distribution of the comparison result of $(x, y_1)$ and $(x, y_2)$ under reward model $r$.
\label{lemma:good_wheel}
\end{lemma}
Putting the above inequalities together, it holds with probability $1-\delta$ that
\begin{align}
    \textbf{Term (i)} &\le -\frac{2}{\alpha}\sum_{t=1}^{T} \sum_{s=1}^{t-1}\exlim{\substack{x^{(s)}\sim \rho,\\ (y_1^{(s)},y_2^{(s)})\sim \pi^{(s)}(\cdot| x^{(s)})}}{D_{\text{H}}^2(\mathbb{P}_{r^{(t)}}(\cdot|x^{(s)},y_1^{(s)},y_2^{(s)})\,\|\,\mathbb{P}_{r^\star}(\cdot|x^{(s)},y_1^{(s)},y_2^{(s)}))} \nonumber \\
    & \qquad + 2 \alpha^{-1} T \log(|\mathcal{R}|/\delta).
    \label{eq:term_i_bound}
\end{align}

\paragraph{Step 2: breaking down term (ii) with the elliptical potential lemma.}
The linear function approximation form \eqref{eq:linear_FA} allows us to write
\begin{align}
    \exlim{x \sim \rho, y \sim \pi_{r_2}(\cdot|x)}{r_1(x, y) - r^\star(x, y) } = \innprod{W(r_1),X(r_2)},
\end{align}
where $X, W : \mathcal{R} \to \mathbb{R}^{d}$ is given by
\begin{equation} \label{eq:def_XW}
    X(r_\theta) = 2C\exlim{x \sim \rho, y \sim \pi_{r_\theta}(\cdot|x)}{\phi(x, y)} , \qquad
    W(r_\theta) = \frac{\theta - \theta^\star}{2C}.
\end{equation}
Let
\begin{equation}
    \Sigma_t = \epsilon I + \sum_{s=1}^{t-1}X(r^{(t)})X(r^{(t)})^{\top}
\end{equation}
for some $\epsilon > 0$.
We begin by decomposing term (ii) as
\begin{align}
  \textbf{Term (ii)}   &= \sum_{t=1}^{T}\exlim{x \sim \rho, y \sim \pi^{(t)}(\cdot|x)}{r^{(t)}(x, y) - r^\star(x, y) } \notag\\
    &= \sum_{t=1}^{T}\innprod{W(r^{(t)}),X(r^{(t)})}\notag\\
    &= \sum_{t=1}^{T}\innprod{W(r^{(t)}),X(r^{(t)})}\mathbf{1}\{\|X(r^{(t)})\|_{\Sigma_t^{-1}}\le 1\}  \nonumber \\
    & \quad\quad\quad + \sum_{t=1}^{T}\innprod{W(r^{(t)}),X(r^{(t)})}\mathbf{1}\{\|X(r^{(t)})\|_{\Sigma_t^{-1}}> 1\},
    \label{eq:online_term_ii_decomp}
\end{align}
where $\mathbf{1}\{A\}$ is an indicator function of event $A$.
To proceed, we recall the elliptical potential lemma for controlling the cumulative sum of $\min\{\|X(r^{(t)})\|_{\Sigma_t^{-1}}^2, 1\}$.
\begin{lemma}[\mbox{\citep[Lemma 11]{abbasi2011improved}}]
Let $\{X_t\}$ be a sequence in $\mathbb{R}^d$ and $\Lambda_0\in\mathbb{R}^{d\times d}$ a positive definite matrix. Define $\Lambda_t = \Lambda_0 + \sum_{s=1}^t X_s X_s^{\top}$. Assume $\|X_t\|\le L$ for all $t$. It holds that
\begin{align*}
    \sum_{t=1}^{T}\min\{\|X_t\|_{\Lambda_t^{-1}}^2, 1\} & \le 2\log \Big(\frac{\text{det}(\Lambda_{T})}{\text{det}(\Lambda_0)}\Big) \\
    &\le 2(d\log((\text{trace}(\Lambda_0)+TL^2)/d) - \log\text{det}(\Lambda_0)).
\end{align*}
\end{lemma}
Applying the above lemma yields
\begin{equation}
    \sum_{t=1}^{T}\min\{\|X(r^{(t)})\|_{\Sigma_t^{-1}}^2, 1\} \le \min\bigg\{2d \log \Big(\frac{4C^4T/d+\epsilon}{\epsilon}\Big), T\bigg\} := d(\epsilon).
\end{equation}

We now control the two terms in \eqref{eq:online_term_ii_decomp}.
\begin{itemize}[leftmargin=16pt]
    \item The first term of \eqref{eq:online_term_ii_decomp} can be bounded by
\begin{align}
    &\sum_{t=1}^{T}\innprod{W(r^{(t)}),X(r^{(t)})}\mathbf{1}\{\|X(r^{(t)})\|_{\Sigma_t^{-1}}\le 1\}\notag\\
    &\le \sum_{t=1}^{T}\|W(r^{(t)})\|_{\Sigma_t}\|X(r^{(t)})\|_{\Sigma_t^{-1}}\mathbf{1}\{\|X(r^{(t)})\|_{\Sigma_t^{-1}}\le 1\} \notag\\
    &\le \sum_{t=1}^{T}\|W(r^{(t)})\|_{\Sigma_t}\min\Big\{\|X(r^{(t)})\|_{\Sigma_t^{-1}},1\Big\}\notag\\
    &= \sum_{t=1}^{T}\Big[\epsilon \|W(r^{(t)})\|_2^2 + \sum_{s=1}^{t-1}\innprod{W(r^{(t)}),X(r^{(s)})}^2\Big]^{1/2}\min\Big\{\|X(r^{(t)})\|_{\Sigma_t^{-1}}^2,1\Big\}^{1/2}\notag\\
    &\overset{\mathrm{(i)}}{\le} \bigg\{\sum_{t=1}^{T}\Big[\epsilon \|W(r^{(t)})\|_2^2 + \sum_{s=1}^{t-1}\innprod{W(r^{(t)}),X(r^{(s)})}^2\Big]\bigg\}^{1/2} \bigg\{\sum_{t=1}^T\min\Big\{\|X(r^{(t)})\|_{\Sigma_t^{-1}}^2,1\Big\}\bigg\}^{1/2}\notag\\
    &\overset{\mathrm{(ii)}}{\le} {\sqrt{d(\epsilon)}}\bigg\{\sum_{t=1}^{T}\sum_{s=1}^{t-1}\innprod{W(r^{(t)}),X(r^{(s)})}^2\bigg\}^{1/2} + {\sqrt{d(\epsilon)\epsilon T}}\notag\\
    &\overset{\mathrm{(iii)}}{\le} \frac{{d(\epsilon)}}{2\mu} + \frac{\mu}{2}\sum_{t=1}^{T}\sum_{s=1}^{t-1}\innprod{W(r^{(t)}),X(r^{(s)})}^2 + {\sqrt{d(\epsilon)\epsilon T}}.
    \label{eq:online_term_ii_decomp_bound_1}
\end{align}
Here, (i) is due to Cauchy–Schwarz inequality, (ii) is due to $\sqrt{a+b} \le \sqrt{a} + \sqrt{b}$ for $\forall a, b \ge 0$, and (iii) results from Young's inequality. We leave the constant $\mu > 0$ to be determined later.
    \item The second term of \eqref{eq:online_term_ii_decomp} can be bounded by 
\begin{align}
    \sum_{t=1}^{T}\innprod{W(r^{(t)}),X(r^{(t)})}\mathbf{1}\{\|X(r^{(t)})\|_{\Sigma_t^{-1}}> 1\} 
    &\le C\sum_{t=1}^{T}\mathbf{1}\{\|X(r^{(t)})\|_{\Sigma_t^{-1}}> 1\}\notag\\
    &\le C\sum_{t=1}^{T}\min\{\|X(r^{(t)})\|_{\Sigma_t^{-1}}^2, 1\} \le Cd(\epsilon),
    \label{eq:online_term_ii_decomp_bound_2}
\end{align}
where the first inequality follows from  $\|X(r^{(t)})\|_2 \le 2C$ and $\|W(r^{(t)})\|_2 \le 1/2$ since $\|\phi(x, y)\|_2\leq 1$. 

\end{itemize}
Putting \eqref{eq:online_term_ii_decomp}, \eqref{eq:online_term_ii_decomp_bound_1} and \eqref{eq:online_term_ii_decomp_bound_2} together, we arrive at
\begin{align}
\textbf{Term (ii)}
    &\le \frac{{d(\epsilon)}}{2\mu} + \frac{\mu}{2}\sum_{t=1}^{T}\sum_{s=1}^{t-1}\innprod{W(r^{(t)}),X(r^{(s)})}^2 + {\sqrt{d(\epsilon)\epsilon T}} + Cd(\epsilon).
    \label{eq:term_ii_tmp}
\end{align}

\paragraph{Step 3: continuing bounding term (ii).} It boils down to control $\innprod{W(r^{(t)}),X(r^{(s)})}^2$. We have
\begin{align}
    \innprod{W(r^{(t)}),X(r^{(s)})}
    &= \exlim{\substack{x \sim \rho,\\ y \sim \pi^{(s)}(\cdot|x)}}{r^{(t)}(x, y) - r^\star(x, y) } \notag\\
    &=\exlim{\substack{x \sim \rho,\\ y_1 \sim \pi^{(s)}(\cdot|x)}}{r^{(t)}(x, y_1) - r^\star(x, y_1) } - \exlim{\substack{x \sim \rho,\\ y_2 \sim \dnorm(\cdot|x)}}{r^{(t)}(x, y_2) - r^\star(x, y_2)} \notag\\
    &=\exlim{\substack{x \sim \rho, \\y_1 \sim \pi^{(s)}(\cdot|x), \\ y_2 \sim \dnorm(\cdot|x)}}{\delta_x(r^{(t)}, r^\star, y_1, y_2)},
    \label{eq:diff_trick}
\end{align}
where $\delta_x(r_1, r_2, y_1, y_2) \coloneqq r_1(x,y_1)-r_1(x,y_2)-(r_2(x,y_1) - r_2(x,y_2))$.
Therefore,
\begin{align}
     \innprod{W(r^{(t)}),X(r^{(s)})}^2 
    &=\exlim{\substack{x \sim \rho, \\y_1 \sim \pi^{(s)}(\cdot|x), \\y_2 \sim \dnorm(\cdot|x)}}{\delta_x(r^{(t)}, r^\star, y_1, y_2)}^2\notag\\
    &= \exlim{\substack{x \sim \rho, \\y_1 \sim \pi^{(s)}(\cdot|x), \\y_2 \sim \dnorm(\cdot|x)}}{\delta_x(r^{(t)}, r^\star, y_1, y_2)^2} - \mathop\text{Var}\limits_{\substack{x \sim \rho, \\y_1 \sim \pi^{(s)}(\cdot|x), \\y_2 \sim \dnorm(\cdot|x)}}[\delta_x(r^{(t)}, r^\star, y_1, y_2)^2]\notag\\
    &\le \exlim{\substack{x \sim \rho, \\y_1 \sim \pi^{(s)}(\cdot|x), \\y_2 \sim \dnorm(\cdot|x)}}{\delta_x(r^{(t)}, r^\star, y_1, y_2)^2}\notag\\
    &\le \sup_{x, y}\frac{\dnorm(y|x)}{\pi^{(s)}(y|x)}\cdot\exlim{\substack{x \sim \rho, \\y_1, y_2 \sim \pi^{(s)}(\cdot|x)}}{\delta_x(r^{(t)}, r^\star, y_1, y_2)^2}\notag\\
    &\le \sup_{x, y}\frac{\pi_{\text{ref}}(y|x)}{\pi^{(s)}(y|x)}\cdot\sup_{x, y}\frac{\dnorm(y|x)}{\pi_{\text{ref}}(y|x)}\cdot \exlim{\substack{x \sim \rho, \\y_1, y_2 \sim \pi^{(s)}(\cdot|x)}}{\delta_x(r^{(t)}, r^\star, y_1, y_2)^2}.
    \label{eq:kappa_dependency}
\end{align}
Recall from \eqref{eq:RLHF-policy-closed-form} that $\pi^{(s)}(y|x) \propto \pi_{\text{ref}}(y|x)\exp(r^{(s)}(x, y)/\beta)$. It follows that $|\log \pi^{(s)}(y|x) - \log \pi_{\text{ref}}(y|x)| \le 2 \|r^{(s)}(x, \cdot)\|_\infty \le 2C/\beta$ (see \eg, \citep[Appendix A.2]{cen2022fast}), and hence $\sup_{x, y}\frac{\pi_{\text{ref}}(y|x)}{\pi^{(s)}(y|x)} \le \exp(2C/\beta)$. To proceed, we demonstrate in the following lemma that $\delta^2$ can be upper bounded by the corresponding Hellinger distance, whose proof is deferred to Appendix~\ref{proof:rhb}.
\begin{lemma}
    \label{lemma:reward_hellinger_bound}
    Assume bounded reward $\|r_1\|_\infty \le C$, $\|r_2\|_\infty \le C$. We have
    \begin{equation*}
        \delta_x(r_1,r_2,y_1,y_2)^2 \le 2(3+\exp(2C))^2 D_{\text{H}}^2(\mathbb{P}_{r_1}(\cdot|x,y_1,y_2)\,\|\,\mathbb{P}_{r_2}(\cdot|x,y_1,y_2)).
    \end{equation*}
\end{lemma}
With the above lemma we arrive at
\begin{align*}
   & \innprod{W(r^{(t)}),X(r^{(s)})}^2 \nonumber \\
   & \le 2(3+\exp(2C))^2\exp(2C/\beta)\kappa\cdot\exlim{\substack{x \sim \rho, \\y_1, y_2 \sim \pi^{(s)}(\cdot|x)}}{D_{\text{H}}^2(\mathbb{P}_{r^{(t)}}(\cdot|x,y_1,y_2)\,\|\,\mathbb{P}_{r^\star}(\cdot|x,y_1,y_2))}.
\end{align*}
where we denote $\kappa = \sup_{x,y}\dfrac{\dnorm(y|x)}{\pi_{\text{ref}}(y|x)}$. Plugging the above bound into \eqref{eq:term_ii_tmp}, we get
\begin{align}
    &\textbf{Term (ii)} \notag\\
    &\le \frac{{d(\epsilon)}}{2\mu} + {\mu}(3+\exp(2C))^2\exp(2C/\beta)\kappa\cdot\sum_{t=1}^{T}\sum_{s=1}^{t-1}\exlim{\substack{x \sim \rho, \\y_1, y_2 \sim \pi^{(s)}(\cdot|x)}}{D_{\text{H}}^2(\mathbb{P}_{r^{(t)}}(\cdot|x,y_1,y_2)\,\|\,\mathbb{P}_{r^\star}(\cdot|x,y_1,y_2))} \notag\\
    &\qquad + 2B{\sqrt{d(\epsilon)\epsilon T}} + Cd(\epsilon).
    \label{eq:term_ii_bound}
\end{align}

\paragraph{Step 4: finishing up.} Combining \eqref{eq:regret_decomp}, \eqref{eq:term_i_bound} and \eqref{eq:term_ii_bound}, with probability $1-\delta$ we have
\begin{align}
    \mathsf{Regret} &\le \frac{2 T \log(|\mathcal{R}|/\delta)}{\alpha} + \frac{{d(\epsilon)}}{2\mu} + {\sqrt{d(\epsilon)\epsilon T}} + Cd(\epsilon)
\end{align}
as long as $\alpha {\mu}(3+\exp(2C))^2\exp(2C/\beta)\kappa \le 2$. Setting $\alpha \asymp \frac{1}{\exp(2C+C/\beta)}\sqrt{\frac{T}{\kappa d(\epsilon)}}$, $\mu \asymp \frac{1}{\exp(2C+C/\beta)}\sqrt{\frac{d(\epsilon)}{\kappa T}}$, and $\epsilon = 1$, we arrive at 
\begin{align*}
    \mathsf{Regret} &\le \widetilde{\mathcal{O}}((\exp(2C+C/\beta))\sqrt{\kappa d T})
\end{align*}
as claimed.


\subsection{Proof of Lemma \ref{lemma:good_wheel}} 
\label{sec:proof_good_wheel}

To begin, we have
\begin{align}
    \ell(r^\star, \mathcal{D}^{(t-1)}) - \ell(r^{(t)}, \mathcal{D}^{(t-1)}) = -\log\frac{\mathbb{P}(\mathcal{D}^{(t-1)}|r^\star)}{\mathbb{P}(\mathcal{D}^{(t-1)}|r^{(t)})} &= -\sum_{s=1}^{t-1}X_{r^{(t)}}^{s},
\end{align}
where we denote
\begin{equation}
    X_r^{s} = \log\frac{\mathbb{P}_{r^\star}(y_+^{(s)}\succ y_-^{(s)}|x^{(s)})}{\mathbb{P}_{r}(y_+^{(s)}\succ y_-^{(s)}|x^{(s)})}.
\end{equation}
To proceed, we recall a useful martingale exponential inequality.
\begin{lemma}[\mbox{\citep[Theorem 13.2]{zhang2023mathematical},\citep[Lemma D.1]{liu2024maximize}}]
Let $\{X_t\}_{t=1}^\infty$ be a sequence of real-valued random variables adapted to filtration $\{\mathcal{F}_t\}_{t=1}^\infty$. It holds with probability $1-\delta$ such that for any $t \ge 1$,
\begin{equation*}
    -\sum_{s=1}^t X_s \le \sum_{s=1}^t \log \ex{}{\exp(-X_s)|\mathcal{F}_{s-1}} + \log (1/\delta).
\end{equation*}
\end{lemma}
Applying the above lemma to $\{\frac{1}{2}X_r^t\}_{t=1}^\infty$ along with the filtration $\{\mathcal{F}_t\}_{t=1}^\infty$ with $\mathcal{F}_t$ given by the $\sigma$-algebra of $\{(x^{(s)}, y_+^{(s)}, y_-^{(s)}): s\le t\}$, we conclude that it holds with probability $1-\delta$ that
\begin{align}
    -\frac{1}{2}\sum_{s=1}^{t-1}X_{r}^{s} &\le \sum_{s=1}^{t-1}\log \ex{}{\exp\Big\{-\frac{1}{2}X_{r}^{s}\Big\}\Big|\mathcal{F}_{s-1}} + \log (|\mathcal{R}|/\delta)\notag\\
    &\le \sum_{s=1}^{t-1}\bigg(\ex{}{\exp\Big\{-\frac{1}{2}X_{r}^{s}\Big\}\Big|\mathcal{F}_{s-1}}-1\bigg) + \log (|\mathcal{R}|/\delta),
    \label{eq:banana}
\end{align}
where the last step results from the inequality $\log(1+x) \le x$ for all $x \ge -1$.
To proceed, note that
\begin{align*}
    &\ex{}{\exp\Big\{-\frac{1}{2}X_{r}^{s}\Big\}\Big|\mathcal{F}_{s-1}} \nonumber \\
    &= \ex{}{\sqrt{\frac{\mathbb{P}_{r}(y_+^{(s)}\succ y_-^{(s)}|x^{(s)})}{\mathbb{P}_{r^\star}(y_+^{(s)}\succ y_-^{(s)}|x^{(s)})}} \,\Big|\, \mathcal{F}_{s-1}}\\
    &= \exlim{\substack{x^{(s)}\sim \rho,\\ (y_1^{(s)},y_2^{(s)})\sim \pi^{(s)}(\cdot|x^{(s)}),\\ (+,-)\sim\mathbb{P}_{r^\star}}}{\sqrt{\frac{\mathbb{P}_{r}(y_+^{(s)}\succ y_-^{(s)}|x^{(s)})}{\mathbb{P}_{r^\star}(y_+^{(s)}\succ y_-^{(s)}|x^{(s)})}}}\\
    &= \exlim{\substack{x^{(s)}\sim \rho,\\ (y_1^{(s)},y_2^{(s)})\sim \pi^{(s)}(\cdot|x^{(s)})}}{\sum_{(+,-)}\sqrt{{\mathbb{P}_{r}(y_+^{(s)}\succ y_-^{(s)}|x^{(s)})\cdot\mathbb{P}_{r^\star}(y_+^{(s)}\succ y_-^{(s)}|x^{(s)})}}}\\
    &= 1 - \frac{1}{2} \exlim{\substack{x^{(s)}\sim \rho,\\ (y_1^{(s)},y_2^{(s)})\sim \pi^{(s)}(\cdot|x^{(s)})}}{\sum_{(+,-)}\Big(\sqrt{\mathbb{P}_{r}(y_+^{(s)}\succ y_-^{(s)}|x^{(s)})}-\sqrt{\mathbb{P}_{r^\star}(y_+^{(s)}\succ y_-^{(s)}|x^{(s)})}\Big)^2}\\
    &=1 - \exlim{\substack{x\sim \rho,\\ (y_1,y_2) \sim \pi^{(s)}(\cdot|x)}}{D_{\text{H}}^2(\mathbb{P}_r(\cdot|x,y_1,y_2\,\|\,\mathbb{P}_{r^\star}(\cdot|x,y_1,y_2)},
\end{align*}
where we denote by $\sum_{(+,-)}$ the summation over different comparison results. Plugging the above equality into \eqref{eq:banana} completes the proof.

\subsection{Proof of Lemma \ref{lemma:reward_hellinger_bound}}
\label{proof:rhb}
By the mean value theorem, we have
\begin{align*}
   \big|\mathbb{P}_{r_1}(y_1\succ y_2|x) - \mathbb{P}_{r_2}(y_1\succ y_2|x)\big| 
    &= \big|\sigma(r_1(x,y_1) - r_1(x,y_2)) - \sigma(r_2(x,y_1) - r_2(x,y_2))\big|\\
    &= \big|\delta_x(r_1,r_2,y_1,y_2) \cdot \sigma'(\xi)\big|\\
    &= \big|\delta_x(r_1,r_2,y_1,y_2)\big| \cdot \sigma(\xi)(1-\sigma(\xi))
\end{align*}
for some $\xi$ between $r_1(x,y_1) - r_1(x,y_2)$ and $r_2(x,y_1) - r_2(x,y_2)$. Since $|\xi| \le 2C$, we have
\begin{equation}
    \sigma(\xi)(1-\sigma(\xi)) \ge \sigma(2C)(1-\sigma(2C)) \ge \frac{1}{3+\exp(2C)}.
\end{equation}
Putting together,
\begin{align*}
    \big|\delta_x(r_1,r_2,y_1,y_2)\big| &\le (3+\exp(2C))\big|\mathbb{P}_{r_1}(y_1\succ y_2|x) - \mathbb{P}_{r_2}(y_1\succ y_2|x)\big|\\
    &= (3+\exp(2C))\text{TV}(\mathbb{P}_{r_1}(\cdot|x,y_1,y_2), \mathbb{P}_{r_2}(\cdot|x,y_1,y_2))\\
    &\le (3+\exp(2C))\sqrt{2}D_{\text{H}}(\mathbb{P}_{r_1}(\cdot|x,y_1,y_2)\,\|\,\mathbb{P}_{r_2}(\cdot|x,y_1,y_2)).
\end{align*}

\section{Analysis for the offline setting}
\label{sec:pf_thm_offline}

\subsection{Proof of Lemma \ref{lemma:saddle_point}}
\label{sec:pf_saddle_point}
By definition, the objective function $\ell({r}, \mathcal{D}) + \alpha J({r}, \pi)$ is strongly concave over $\pi$, and convex over $r$. By Danskin's theorem, we have 
\begin{align*}
    &\nabla_r\big(\max_\pi [\ell(\widehat{r}, \mathcal{D}) + \alpha J(\widehat{r}, \pi)]\big)= \nabla_r\big(\ell(\widehat{r}, \mathcal{D}) + \alpha J(\widehat{r}, \widehat{\pi})\big).
\end{align*}
Therefore, for any $r'$, by convexity of the objective function we have
\begin{align*}
     \ell({r}', \mathcal{D}) + \alpha J({r}', \widehat{\pi}) &\ge \ell(\widehat{r}, \mathcal{D}) + \alpha J(\widehat{r}, \widehat{\pi}) + \innprod{r' - \widehat{r}, \nabla_r\big(\ell(\widehat{r}, \mathcal{D}) + \alpha J(\widehat{r}, \widehat{\pi})\big)}\\
     &=\ell(\widehat{r}, \mathcal{D}) + \alpha J(\widehat{r}, \widehat{\pi}) + \innprod{r' - \widehat{r}, \nabla_r\big(\max_\pi [\ell(\widehat{r}, \mathcal{D}) + \alpha J(\widehat{r}, \pi)]\big)}\\
     &\ge \ell(\widehat{r}, \mathcal{D}) + \alpha J(\widehat{r}, \widehat{\pi}).
\end{align*}
The last line is due to the definition of $\widehat{r}$ (c.f.~\eqref{eq:pessimism_r}). The other relation, 
$\ell(\widehat{r}, \mathcal{D}) + \alpha J(\widehat{r}, \widehat{\pi}) \ge \ell(\widehat{r}, \mathcal{D}) + \alpha J(\widehat{r}, \pi')$, follows directly from the definition of $\widehat{\pi}$ (c.f.~\eqref{eq:RBMLE_offline}).

\subsection{Proof of Theorem~\ref{thm:offline}}
We  decompose the sub-optimality gap of $\widehat{\pi}$ by
\begin{align}
    &J^\star(r^\star) - J(r^\star, \widehat{\pi}) \notag\\
    &= \big[J(r^\star, \pi^\star) - J(\widehat{r}, \pi^\star)\big] + \big[J(\widehat{r}, \pi^\star) - J(\widehat{r}, \widehat{\pi})\big] + \big[J(\widehat{r}, \widehat{\pi})- J(r^\star, \widehat{\pi})\big]\notag\\
    &\le \underbrace{\big[J(r^\star, \pi^\star) - J(\widehat{r}, \pi^\star)\big]}_{\textbf{Term (i)}} + \underbrace{\big[J(\widehat{r}, \widehat{\pi})- J(r^\star, \widehat{\pi})\big]}_{\textbf{Term (ii)}},
    \label{eq:offline_decomp}
\end{align}
where the last line is due to $J(\widehat{r}, \pi^\star) \leq J(\widehat{r}, \widehat{\pi})$ according to the definition of $\widehat{\pi}$ (c.f.~\eqref{eq:RBMLE_offline}). We proceed to bound the two terms separately. Here we have written $\widehat{r} = r_{\widehat{\theta}}$ for notational simplicity. In addition, we denote the MLE estimate by $r_{\mathsf{MLE}} = r_{\theta_{\mathsf{MLE}}}$.

By the definition of $J(r,\pi)$ (cf.~\eqref{eq:KL_reward}), it follows that term (i) in \eqref{eq:offline_decomp} can be further decomposed as
\begin{align}
\textbf{Term (i)} &= \exlim{\substack{x \sim \rho,\\y\sim\pi^\star(\cdot|x)}}{r^\star(x,y) - \widehat{r}(x,y)} \nonumber \\
    &= \exlim{\substack{x \sim \rho,\\y\sim\pi^\star(\cdot|x)}}{\innprod{\phi(x,y), \theta^\star - \widehat{\theta}}}\nonumber \\
    &= \underbrace{ \exlim{\substack{x \sim \rho,\\y\sim\pi^\star(\cdot|x)}}{\innprod{\phi(x,y), \theta^\star - {\theta}_{\mathsf{MLE}}}} }_{\textbf{Term (ia)}}+  \underbrace{ \exlim{\substack{x \sim \rho,\\y\sim\pi^\star(\cdot|x)}}{\innprod{\phi(x,y), \theta_{\mathsf{MLE}} - \widehat{\theta}}} }_{\textbf{Term (ib)}}, \label{eq:apple}
\end{align}
where $r_{\mathsf{MLE}}(x,y) = \langle \phi(x,y) , \theta_{\mathsf{MLE}} \rangle$.

\paragraph{Step 1: bounding term (ia).} 
To continue, we recall a useful lemma from \citep{zhu2023principled}.
\begin{lemma}[\mbox{\citep[Lemma 3.1]{zhu2023principled}}]
    \label{lemma:zhu2023}
    For any $\lambda > 0$ and $\delta\in (0,1)$, with probability at least $1-\delta$,
    \begin{equation*}
        \|\theta_{\mathsf{MLE}} - \theta^\star\|_{\Sigma_{\mathcal{D}}+\lambda I} \le \mathcal{O}\Bigg((3+\exp(C))\sqrt{{\frac{d + \log (1/\delta)}{N}} }+ \sqrt{\lambda C^2}\Bigg).
    \end{equation*}
    In addition, we have
    \begin{equation} \label{eq:scvx}
        \frac{1}{3+\exp(C)}\Sigma_{\mathcal{D}} \preceq \frac{1}{N} \nabla_\theta^2 \ell(r_\theta, \mathcal{D}) \preceq \frac{1}{4}\Sigma_{\mathcal{D}}
    \end{equation}
    for all $\theta$ such that $\|r_\theta\|_\infty \le C.$
\end{lemma}

The first term of \eqref{eq:apple} can be bounded with Lemma \ref{lemma:zhu2023} as
\begin{align}
  \textbf{Term (ia)}   &\le \|\theta^\star - {\theta}_{\mathsf{MLE}}\|_{\Sigma_\mathcal{D} + \lambda I}\cdot\Big\|\exlim{\substack{x \sim \rho,\\y\sim\pi^\star(\cdot|x)}}{\phi(x, y)}\Big\|_{(\Sigma_\mathcal{D} + \lambda I)^{-1}} \nonumber \\
    &\le \mathcal{O}\Bigg(\Big((3+\exp(C))\sqrt{{\frac{d + \log (1/\delta)}{N}} }+ \sqrt{\lambda C^2}\Big)\cdot\Big\|\exlim{\substack{x \sim \rho,\\y\sim\pi^\star(\cdot|x)}}{\phi(x, y)}\Big\|_{(\Sigma_\mathcal{D} + \lambda I)^{-1}}\Bigg). \label{eq:bound_ia}
\end{align}

\paragraph{Step 2: bounding term (ib).} For the second term of \eqref{eq:apple}, recall that
\begin{align*}
    \widehat{r} = \arg\min_{r\in\mathcal{R}} \big\{\ell({r}, \mathcal{D}) + \alpha J({r}, \widehat{\pi}) \big\},
\end{align*}
or equivalently
\begin{align*}
    \widehat{\theta} = \arg\min_{\theta\in\Theta} \big\{\ell({r}_\theta, \mathcal{D}) + \alpha J({r}_\theta, \widehat{\pi}) \big\},
\end{align*}
and that
\begin{align*}
    {\theta}_{\mathsf{MLE}} = \arg\min_{\theta\in\Theta} \ell({r}_\theta, \mathcal{D}).
\end{align*}
With linear constraint \eqref{eq:THETA}, by KKT condition we have
\begin{equation*}
    \nabla_\theta \ell(\widehat{r}, \mathcal{D}) + \alpha \exlim{\substack{x\sim\rho,\\ y \sim \widehat{\pi}(\cdot| x)}}{\phi(x, y)} + \lambda_1 \exlim{\substack{x\sim\rho,\\ y \sim \dnorm(\cdot| x)}}{\phi(x, y)} = 0
\end{equation*}
for some $\lambda_1 \in \mathbb{R}$, and
\begin{equation*}
    \nabla_\theta \ell({r}_{\mathsf{MLE}}, \mathcal{D}) + \lambda_2 \exlim{\substack{x\sim\rho,\\ y \sim \dnorm(\cdot| x)}}{\phi(x, y)} = 0
\end{equation*}
for some $\lambda_2 \in \mathbb{R}$. By strong monotonicity of $\nabla_\theta\ell$ (cf.~\eqref{eq:scvx}), we have
\begin{align*}
    \frac{N}{3+\exp(C)}\big\|\widehat{\theta} - \theta_{\mathsf{MLE}}\big\|_{\Sigma_\mathcal{D}}^2      &\le \innprod{\nabla_\theta {\ell}(\widehat{r}, \mathcal{D}) - \nabla_\theta {\ell}({r}_{\mathsf{MLE}}, \mathcal{D}), \widehat{\theta} - \theta_{\mathsf{MLE}}}\notag\\
    &=\Big\langle{-\alpha \exlim{\substack{x\sim\rho,\\ y \sim \widehat{\pi}(\cdot| x)}}{\phi(x, y)} - (\lambda_1 -\lambda_2) \exlim{\substack{x\sim\rho,\\ y \sim \dnorm(\cdot| x)}}{\phi(x, y)}, \widehat{\theta} - \theta_{\mathsf{MLE}}}\Big\rangle\notag\\
    &=-\alpha\Big\langle{\exlim{\substack{x\sim\rho,\\ y \sim \widehat{\pi}(\cdot| x)}}{\phi(x, y)} - \exlim{\substack{x\sim\rho,\\ y \sim \dnorm(\cdot| x)}}{\phi(x, y)}, \widehat{\theta} - \theta_{\mathsf{MLE}}}\Big\rangle\notag\\
    &\le\alpha\Big\|{\exlim{\substack{x\sim\rho,\\ y \sim \widehat{\pi}(\cdot| x)}}{\phi(x, y)} - \exlim{\substack{x\sim\rho,\\ y \sim \dnorm(\cdot| x)}}{\phi(x, y)}\Big\|_{(\Sigma_{\mathcal{D}}+\lambda I)^{-1}}\big\|\widehat{\theta} - \theta_{\mathsf{MLE}}}\big\|_{\Sigma_{\mathcal{D}}+\lambda I}\notag\\
    &\le \alpha \kappa_{\mathcal{D}}\big\| \widehat{\theta} - \theta_{\mathsf{MLE}}\big\|_{\Sigma_\mathcal{D} + \lambda I},
\end{align*}
where we denote
\begin{equation}\label{eq:def_kappa_offline}
\kappa_{\mathcal{D}} = \Big\|\exlim{\substack{x\sim\rho,\\ y \sim \widehat{\pi}(\cdot| x)}}{\phi(x, y)} - \exlim{\substack{x\sim\rho,\\ y \sim \dnorm(\cdot| x)}}{\phi(x, y)}\Big\|_{(\Sigma_\mathcal{D} + \lambda I)^{-1}}.
\end{equation}

The penultimate step results from $\widehat{\theta}, \theta_{\mathsf{MLE}}\in\Theta$, which ensures 
\begin{equation*}
    \Big\langle\exlim{\substack{x\sim\rho,\\ y \sim \dnorm(\cdot| x)}}{\phi(x, y)}, \widehat{\theta}\Big\rangle = \Big\langle\exlim{\substack{x\sim\rho,\\ y \sim \dnorm(\cdot| x)}}{\phi(x, y)}, {\theta}_{\mathsf{MLE}}\Big\rangle = 0 
\end{equation*}
It follows that
\begin{align*}
    \frac{N}{3+\exp(C)}\big\|\widehat{\theta} - \theta_{\mathsf{MLE}}\big\|_{\Sigma_\mathcal{D}+\lambda I}^2
    &\le \frac{N}{3+\exp(C)}\big\|\widehat{\theta} - \theta_{\mathsf{MLE}}\big\|_{\Sigma_\mathcal{D}}^2 +  \frac{N}{3+\exp(C)}\big\|\widehat{\theta} - \theta_{\mathsf{MLE}}\big\|_{\lambda I}^2\\
    &\le \alpha \kappa_{\mathcal{D}}\big\|\widehat{\theta} - \theta_{\mathsf{MLE}}\big\|_{\Sigma_\mathcal{D} + \lambda I} + \frac{N\lambda C^2}{3+\exp(C)}.
\end{align*}
The above inequality allows us to bound
\begin{equation}
    \big\|\widehat{\theta} - \theta_{\mathsf{MLE}}\big\|_{\Sigma_\mathcal{D} + \lambda I} \le \frac{\alpha(3+\exp(C))}{N}\kappa_{\mathcal{D}} + 2\sqrt{\lambda C^2}.
    \label{eq:ugly_bound}
\end{equation}
Therefore, the second term of \eqref{eq:apple} can be bounded as
\begin{align}
  \textbf{Term (ib)}  &\le \big\|\widehat{\theta} - \theta_{\mathsf{MLE}}\big\|_{\Sigma_\mathcal{D} + \lambda I} \Big\|\exlim{\substack{x \sim \rho,\\y\sim\pi^\star(\cdot|x)}}{\phi(x, y)}\Big\|_{(\Sigma_\mathcal{D} + \lambda I)^{-1}}\nonumber\\
    &\le \bigg(\frac{\alpha(3+\exp(C))}{N}\kappa_{\mathcal{D}} + 2\sqrt{\lambda C^2}\bigg)\Big\|\exlim{\substack{x \sim \rho,\\y\sim\pi^\star(\cdot|x)}}{\phi(x, y)}\Big\|_{(\Sigma_\mathcal{D} + \lambda I)^{-1}}. \label{eq:bound_ib}
\end{align}
Putting \eqref{eq:bound_ia} and \eqref{eq:bound_ib} together, we have
\begin{align}
    \textbf{Term (i)} 
    &\le \mathcal{O}\Bigg(\bigg[\frac{3+\exp(C)}{\sqrt{N}}\Big(\sqrt{{{d + \log (1/\delta)}}} + \frac{\alpha}{\sqrt{N}}\kappa_{\mathcal{D}}\Big)+ \sqrt{\lambda C^2}\bigg]  \cdot \Big\|\exlim{\substack{x \sim \rho,\\y\sim\pi^\star(\cdot|x)}}{\phi(x, y)}\Big\|_{(\Sigma_\mathcal{D} + \lambda I)^{-1}}\Bigg).
    \label{eq:uugly_bound_1}
\end{align}

\paragraph{Step 3: bounding term (ii).}
We can decompose and bound term (ii) by
\begin{align*}
    J(\widehat{r}, \widehat{\pi})- J(r^\star, \widehat{\pi}) &= J(\widehat{r}, \widehat{\pi}) + \frac{1}{\alpha}\ell(\widehat{r}, \mathcal{D}) - \Big(J(r^\star, \widehat{\pi}) + \frac{1}{\alpha}\ell(r^\star, \mathcal{D})\Big) + \frac{1}{\alpha}(\ell(\widehat{r}, \mathcal{D}) - \ell({r}^\star, \mathcal{D})) \nonumber \\
    &\overset{(i)}{\le} \frac{1}{\alpha}(\ell(\widehat{r}, \mathcal{D}) - \ell({r}^\star, \mathcal{D}))\\
    &\le \frac{1}{\alpha}(\ell(\widehat{r}, \mathcal{D}) - \ell({r}_{\mathsf{MLE}}, \mathcal{D}) + \ell({r}_{\mathsf{MLE}}, \mathcal{D}) - \ell({r}^\star, \mathcal{D})),
\end{align*}
where (i) follows from the fact that $(\widehat{r}, \widehat{\pi})$ is a saddle point.
Due to convexity of $\ell$, we have
\begin{align*}
    \ell(\widehat{r}, \mathcal{D}) - \ell({r}_{\mathsf{MLE}}, \mathcal{D}) &\le \innprod{\nabla_\theta {\ell}(\widehat{r}, \mathcal{D}), \widehat{\theta} - \theta_{\mathsf{MLE}}} \\
    &=\innprod{-\alpha\exlim{\substack{x\sim\rho,\\ y \sim \widehat{\pi}(\cdot| x)}}{\phi(x, y)} - \lambda_1 \exlim{\substack{x\sim\rho,\\ y \sim \dnorm(\cdot| x)}}{\phi(x, y)}, \widehat{\theta} - \theta_{\mathsf{MLE}}}\\
    &=-\alpha\innprod{\exlim{\substack{x\sim\rho,\\ y \sim \widehat{\pi}(\cdot| x)}}{\phi(x, y)} - \exlim{\substack{x\sim\rho,\\ y \sim \dnorm(\cdot| x)}}{\phi(x, y)}, \widehat{\theta} - \theta_{\mathsf{MLE}}}\\
    &\le \alpha \kappa_{\mathcal{D}}\| \widehat{\theta} - \theta_{\mathsf{MLE}}\|_{\Sigma_\mathcal{D} + \lambda I}\\
    &\le \frac{\alpha^2(3+\exp(C))}{N}\kappa_{\mathcal{D}}^2 + 2\sqrt{\lambda C^2}\alpha \kappa_{\mathcal{D}},
\end{align*}
where the last step is due to \eqref{eq:ugly_bound}. On the other hand, with probability $1-\delta$ we have \citep[Lemma 1]{zhan2023provable}:
\begin{equation*}
    \ell({r}_{\mathsf{MLE}}, \mathcal{D}) - \ell({r}^\star, \mathcal{D}) \le \widetilde{\mathcal{O}}(1).
\end{equation*}
Putting pieces together, 
\begin{align}
    &\textbf{Term (ii)} \le \frac{\alpha(3+\exp(C))}{N}\kappa_{\mathcal{D}}^2 + 2\sqrt{\lambda C^2} \kappa_{\mathcal{D}} + \frac{1}{\alpha}.
    \label{eq:uugly_bound_2}
\end{align}

\paragraph{Step 4: putting things together.} Combining \eqref{eq:offline_decomp} \eqref{eq:uugly_bound_1}, \eqref{eq:uugly_bound_2}, with probability $1-\delta$ we have
\begin{align*}
    &J^\star(r^\star) - J(r^\star, \widehat{\pi}) \\
    &\le \mathcal{O}\Bigg(\frac{1}{\sqrt{N}}\bigg[{(3+\exp(C))}\Big(\sqrt{{{d + \log (1/\delta)}}} + \kappa_{\mathcal{D}}\Big)+ C\bigg] \cdot\Big\|\exlim{\substack{x \sim \rho,\\y\sim\pi^\star(\cdot|x)}}{\phi(x, y)}\Big\|_{(\Sigma_\mathcal{D} + \lambda I)^{-1}}\\
    &\qquad + \frac{1}{\sqrt{N}}\Big({(3+\exp(C))}\kappa_{\mathcal{D}}^2 + 2C\kappa_{\mathcal{D}}  + 1\Big)\Bigg).
\end{align*}
Here we have set $\alpha = \sqrt{N}$ and $\lambda = 1/N$.
We conclude by bounding $\kappa_{\mathcal{D}}$ as
\begin{align*}
    \kappa_{\mathcal{D}}^2 &= \Big\|{\exlim{\substack{x\sim\rho,\\ y \sim \widehat{\pi}(\cdot| x)}}{\phi(x, y)} - \exlim{\substack{x\sim\rho,\\ y \sim \dnorm(\cdot| x)}}{\phi(x, y)}\Big\|^2_{(\Sigma_{\mathcal{D}}+\lambda I)^{-1}}}\\
    &\le \Big\|{\exlim{\substack{x\sim\rho,\\ y \sim \widehat{\pi}(\cdot| x)}}{\phi(x, y)} - \exlim{\substack{x\sim\rho,\\ y \sim \dnorm(\cdot| x)}}{\phi(x, y)}\Big\|^2_{2}} \cdot \Big\|(\Sigma_{\mathcal{D}}+\lambda I)^{-1}\Big\|_2\\
    &\le 4 (\lambda_{\min}(\Sigma_\mathcal{D}) + \lambda)^{-1} . 
\end{align*}

\section{Experimental details}\label{appendix:exp_details}
\subsection{RLHF for LLMs --- offline setting}
For the offline setting experiments, we adopt instruction tuned models, \textsc{Llama2-7b-chat}, \textsc{Llama2-13b-chat} and \textsc{Flan-T5-xl} as the base models.
To prompt these models, we prepend the questions in \texttt{ARC-Challenge} task~\citep{clark2018think} with 
\begin{center}
\textit{What is the choice to the following Question? Only provide the choice by providing a single letter.} \end{center}
and further append the question with 
\begin{center}
\textit{The answer is:}.    
\end{center}
The question is structured in a way that the multiple choices are appended with alphabets (letters) within parenthesis to the question. As an example:

\textit{Question: George wants to warm his hands quickly by rubbing them. Which skin surface will produce the most heat? Choices: (A)dry palms (B)wet palms (C)palms covered with oil (D)palms covered with lotion}

We set $\dnorm$ to the empirical distribution of the ground truth answer which is known to us.
Based on preliminary experiments, we set $\beta$ as 0.1 in DPO and $\tau$ as 1.0 in IPO.
For VPO, we experiment with moving $\alpha$ from 0.01 to 10, choosing 1 for the reported results.
For all models, we train the base models with different algorithms DPO, VPO and IPO for 3000 steps and report the accuracy of the performance on the \texttt{ARC-challenge} test data set after every 500 steps. 
The training for \textsc{Llama2-13b-chat} model on 128 TPU-v4 takes around 2hrs and for \textsc{Flan-T5-xl} on 64 TPU-v3 takes 1 hour.

\subsection{RLHF for LLMs --- online setting}
\paragraph{Buffer.}
The prompt used for generating AI feedback (and rating) for \texttt{TL;DR} summarization is identical to ~\citep{guo2024direct}.
We set $\dnorm$ to the empirical distribution of the negative answer pairs $(x, y_-)$ collected by the policy. We set $\beta$ as 0.1 for the DPO term similar to ~\citep{guo2024direct}. Additionally for VPO, we decrease the coefficient exponentially following $\frac{\alpha}{\sqrt{1+\text{training steps}}}$. We try different values of $\alpha$ and report the results for 0.1 and 0.01.

The training of the policy, \textsc{PaLM2-XXS} on 64 TPU-v3 for 5000 steps takes around 12 hours for both online DPO and VPO.
We report the win rate percentage against the base SFT model for every 1000 steps using \textsc{PaLM2-XS} judge.
We also further conduct side by side comparison of Online DPO and VPO at 5000 step.

\paragraph{Iterative.}
The UltraFeedback data \citep{cui2023ultrafeedback} contains around 61k preference pairs of single-turn conversations.
We divide this data set into 4 chunks.
We use the first chunk to train a DPO model which we refer to as Zephyr-7B-DPO in Table~\ref{tab:models}.
We use the remaining 3 chunks to train consecutive iterations of DPO and VPO, using the checkpoint from the previous iteration to initialize the policy for the current iteration.
For each iteration, the prompts from the data are extracted and a new answer is sampled from the policy.
We label the data using online AI feedback provided by PairRM \citep{jiang2023llm}, using the similar ranking procedure as \citep{zhang2024self} where the new sample is ranked against $y_w$ and $y_l$ from the data.
For VPO, we also adopt another ranking process where we sample two answer from the policy which are ranked against each other.
We also set $\dnorm$ to the empirical distribution of the negative answer pairs $(x, y_-)$ collected by the policy.
We report the results of the best checkpoint of these two ranking procedures in Table~\ref{tab:models}. All experiments are conducted on 16xA100 GPUs.

\begin{figure*}[h]
\centering
\includegraphics[width=0.6\linewidth]{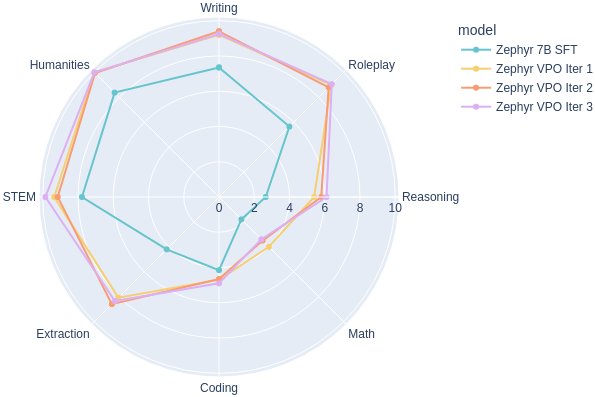}
\caption{
Radar chart of MT-Bench results for Zephyr-7B.}
\label{fig:radar_chart_mt_bench}
\end{figure*}

\end{document}